\documentclass{article}

\usepackage{arxiv}
\usepackage{graphicx}
\graphicspath{{Figures/},.}
\usepackage{amsfonts, bm, amsmath, amssymb}
\usepackage{mathtools}
\usepackage[table]{xcolor}
\usepackage{xcolor}
\usepackage{tabularx}
\usepackage{pdflscape}
\usepackage{multirow}

\definecolor{myred}{RGB}{153, 0, 0}
\definecolor{highlight}{RGB}{254, 232, 200}

\usepackage{makecell}
\usepackage{setspace}

\usepackage{comment}
\usepackage[utf8]{inputenc} 
\usepackage[T1]{fontenc}    
\usepackage{hyperref}       
\usepackage{url}            
\usepackage{booktabs}       
\usepackage{nicefrac}       
\usepackage{microtype}      
\usepackage[numbers,sort&compress]{natbib}

\usepackage{doi}
\usepackage{algorithm}
\usepackage{algorithmicx}
\usepackage{algpseudocode}%
\newcommand*{\vertbar}{\rule[-1ex]{0.5pt}{2.5ex}}

\algrenewcommand\algorithmicrequire{\textbf{Input:}}
\algrenewcommand\algorithmicensure{\textbf{Output:}}
\algnewcommand{\LeftComment}[1]{\Statex \(\triangleright\) #1}
\DeclareMathOperator*{\argmin}{\arg\min}

\newcommand{\reals}{\mathbb{R}}
\newcommand{\abs}[1]{\vert#1\vert}

\title{Automatically identifying ordinary differential equations from data}
\date{} 					

\author{ 
    \href{https://orcid.org/0000-0001-9748-8687}{\includegraphics[scale=0.06]{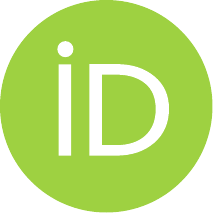}\hspace{1mm}Kevin Egan} \\
	Department of Engineering\\
        Durham University\\
	Durham, UK, DH1 3HN\\
	\texttt{kevin.egan@dur.ac.uk} \\
	\And
	\href{https://orcid.org/0000-0001-9368-236X}{\includegraphics[scale=0.06]{orcid.pdf}\hspace{1mm}Weizhen Li} \\
	Department of Engineering\\
        Durham University\\
	Durham, UK, DH1 3HN\\
	\texttt{weizhen.li@dur.ac.uk} \\
    \And
    \href{https://orcid.org/0000-0002-3279-4218}{\includegraphics[scale=0.06]{orcid.pdf}\hspace{1mm}Rui Carvalho} \\
	Department of Engineering\\
        Durham University\\
	Durham, UK, DH1 3HN\\
	\texttt{rui.carvalho@dur.ac.uk} \\
}

\renewcommand{\shorttitle}{\textit{arXiv} Template}
\hypersetup{
pdftitle={A template for the arxiv style},
pdfsubject={q-bio.NC, q-bio.QM},
pdfauthor={David S.~Hippocampus, Elias D.~Striatum},
pdfkeywords={First keyword, Second keyword, More},
}

\begin{document}
\maketitle
\begin{abstract}
Discovering nonlinear differential equations that describe system dynamics from empirical data is a fundamental challenge in contemporary science.
Here, we propose a methodology to identify dynamical laws by integrating denoising techniques to smooth the signal, sparse regression to identify the relevant parameters, and bootstrap confidence intervals to quantify the uncertainty of the estimates.
We evaluate our method on well-known ordinary differential equations with an ensemble of random initial conditions, time series of increasing length, and varying signal-to-noise ratios.
Our algorithm consistently identifies three-dimensional systems, given moderately-sized time series and high levels of signal quality relative to background noise.
By accurately discovering dynamical systems automatically, our methodology has the potential to impact the understanding of complex systems, especially in fields where data are abundant, but developing mathematical models demands considerable effort.
\end{abstract}


\section{Introduction}
\label{sect:intro}
Since Newton discovered the second law of motion, scientists have sought to formulate mathematical models in the form of differential equations that accurately represent natural phenomena.
In the past half-century, dynamical systems have been employed in various disciplines such as physics~\cite{ruelleNatureTurbulence1971, wattsCollectiveDynamicsSmallworld1998}, chemistry~\cite{petrovControllingChaosBelousov1993}, biology~\cite{mackeyOscillationChaosPhysiological1977,tysonNetworkDynamicsCell2001,steuerStructuralKineticModeling2006,karsentiSelforganizationCellBiology2008,kholodenkoSignallingBalletSpace2010,altrockMathematicsCancerIntegrating2015}, neuroscience~\cite{chialvoEmergentComplexNeural2010,todorovOptimalityPrinciplesSensorimotor2004,breakspearDynamicModelsLargescale2017}, epidemiology~\cite{sugiharaNonlinearForecastingWay1990,earnSimpleModelComplex2000},  ecology~\cite{sugiharaDetectingCausalityComplex2012,woodStatisticalInferenceNoisy2010} and environmental sciences~\cite{nicolisReconstructionDynamicsClimatic1986, steffenTrajectoriesEarthSystem2018}.
Nonetheless, developing these models remains challenging and typically requires considerable effort from specialists in the relevant fields~\cite{waltzAutomatingScience2009,schmidtAutomatedRefinementInference2011}.

As early as the 1980s, scientists turned to statistical methods to reverse engineer governing equations for nonlinear systems from data, aiming to automatically discover mathematical models that accurately represent the inherent dynamics~\cite{crutchfieldEquationsMotionData1987}.
This approach is often referred to as the inverse problem~\cite{tarantolaInverseProblemTheory2005} or system identification~\cite{hongModelSelectionApproaches2008}.
Subsequent advances have led to significant progress in identifying differential equations symbolically~\cite{schmidtDistillingFreeFormNatural2009,schaefferSparseDynamicsPartial2013,danielsAutomatedAdaptiveInference2015,bruntonDiscoveringGoverningEquations2016,raissiMachineLearningLinear2017,schaefferLearningPartialDifferential2017,schaefferExtractingSparseHighDimensional2018,raissiHiddenPhysicsModels2018,luschDeepLearningUniversal2018,zhangRobustDatadrivenDiscovery2018,guimeraBayesianMachineScientist2020,udrescuAIFeynmanPhysicsinspired2020,cortiellaSparseIdentificationNonlinear2021,callahamLearningDominantPhysical2021}.
One practical method that has emerged from this effort is sparse regression, which eliminates the time-consuming task of determining equations manually~\cite{yuanDataDrivenDiscovery2019}.
A remarkable breakthrough is the sparse identification of nonlinear dynamics (SINDy)~\cite{bruntonDiscoveringGoverningEquations2016}, an approach that employs a sparsity-promoting framework to identify interpretable models from data by only selecting the most dominant candidate terms from a high-dimensional nonlinear-function space. 

SINDy with AIC, an extension of the original SINDy algorithm, uses a grid of sparsity-promoting threshold parameters in conjunction with the Akaike information criterion (AIC), a statistical method for model comparison, to determine the model that most accurately characterizes the dynamics of a given system~\cite{manganModelSelectionDynamical2017,bruntonDiscoveringGoverningEquations2016}.
Although this method automates model selection, it encounters several obstacles that limit its practicality.
Key challenges include its dependence on prior knowledge of the governing equations for model validation and identification, as well as the requirement for high-quality measurements given its limited capacity to compute optimal numerical derivatives from data.
Furthermore, the efficacy of SINDy with AIC has only been demonstrated on data sets generated using specific initial conditions, sufficient observations, and low levels of noise, indicating the need for more comprehensive and rigorous analyses to assess its performance in diverse settings.

Here, we present an algorithm for automatically identifying dynamical systems from data by optimally calculating derivatives and incorporating formal variable selection procedures for model inference and selection.
Our method tolerates low to medium levels of noise, requiring only the assumptions of model sparsity and the presence of the governing terms in the design matrix.
To demonstrate the effectiveness of our approach, we examine its success rate on synthetic data sets generated from known ordinary differential equations, exploring a range of initial conditions, time series of increasing length, and various noise intensities.
Our algorithm automates the discovery of three-dimensional systems more efficiently than SINDy with AIC, achieving higher identification accuracy with moderately-sized data sets and high signal quality.

\section{Results}
\label{sect:results}

\subsection{Modeling systems of ODEs with linear regression}
\label{subsect:dd_system_id}
Ordinary differential equations (ODEs) are often used to model dynamical systems in the form of
\begin{equation}
\label{dynamical_systems}
    \frac{d}{dt}x_j(t) = \dot{x}_j(t) = f_j(x(t)) \qquad j = 1,\dotsc,m,
\end{equation}
where $x = x(t)$ = $\left(x_{1}(t)\ \ x_{2}(t) \ \cdots \ \ x_{m}(t)\right)^T \in \mathbb{R}^{m}$ is a state space vector, and $f(x(t)): \mathbb{R}^{m} \rightarrow \mathbb{R}^{m}$ describes the system's evolution in time~\cite{guckenheimerNonlinearOscillationsDynamical2013}.
We approximate the dynamics symbolically by
\begin{align}
\label{glm_dynamical_systems}
    \dot{x}_j \approx \theta_{\text{F}}^T(x)\beta_j,  \qquad j = 1,\dotsc,m,
\end{align}
where $\beta_j \in \mathbb{R}^{p}$ is a sparse coefficient vector of system parameters and $\theta_{\text{F}}(x)$ is a feature vector containing $p$ symbolic functions, each representing an ansatz that we can use to describe the dynamics.

To identify the system from data, we first construct a state matrix $\mathbf{\Tilde{X}}\in \reals^{n\times m}$ from measurements of $x(t)$ taken at times $t_{1},t_{2},...,t_{n}$, then apply the Savitzky-Golay filter~\cite{savitzkySmoothingDifferentiationData1964} to smooth each column $\mathbf{x}_j=SG(\mathbf{\Tilde{x}}_j)$ and calculate the derivative $\mathbf{\dot{x}}_j$. We next consolidate $\mathbf{X}$ and $\mathbf{\dot{X}}$ and build the block design matrix ${\mathbf{\Theta}(\mathbf{X})\in \reals^{n\times p}}$:
\begin{equation}
\label{theta_matrix}
\mathbf{\Theta}(\mathbf{X}) = 
\begin{pmatrix}

    \vertbar & \vertbar & \vertbar &        & \vertbar & \vertbar\\
    \mathbf{1}    & \mathbf{X}    & \mathbf{X}^{[2]} & \cdots & \mathbf{X}^{[d]} & \mathbf{\Phi}(\mathbf{X})\\
    \vertbar & \vertbar & \vertbar &       & \vertbar & \vertbar

\end{pmatrix} ,
\end{equation}
where $\mathbf{X}^{[i]}$ for $i = 1,\dotsc, d$ is a matrix whose column vectors denote all  monomials of order $i$ in $x(t)$, and $\mathbf{\Phi}(\mathbf{X})$ can contain nonlinear functions such as trigonometric, logarithmic, or exponential~\cite{bruntonDiscoveringGoverningEquations2016}.

Finally, we perform a linear regression with the above matrices:
\begin{equation}
\label{derivative_system_id_equation_with_noise}
    \mathbf{\dot{X}} = \mathbf{\Theta}(\mathbf{X}){\mathbf{B}} + \mathbf{E},
\end{equation}
where $\mathbf{B}\in \reals^{p\times m}$ and $\mathbf{E}\in\reals^{n\times m}$ denote the coefficient and residual matrices, respectively.

\subsection{Automatic regression for governing equations (ARGOS)}
\label{subsect:auto_sparse_reg_description}
\begin{figure*}[!htbp]
\centering
\includegraphics[width=\textwidth]{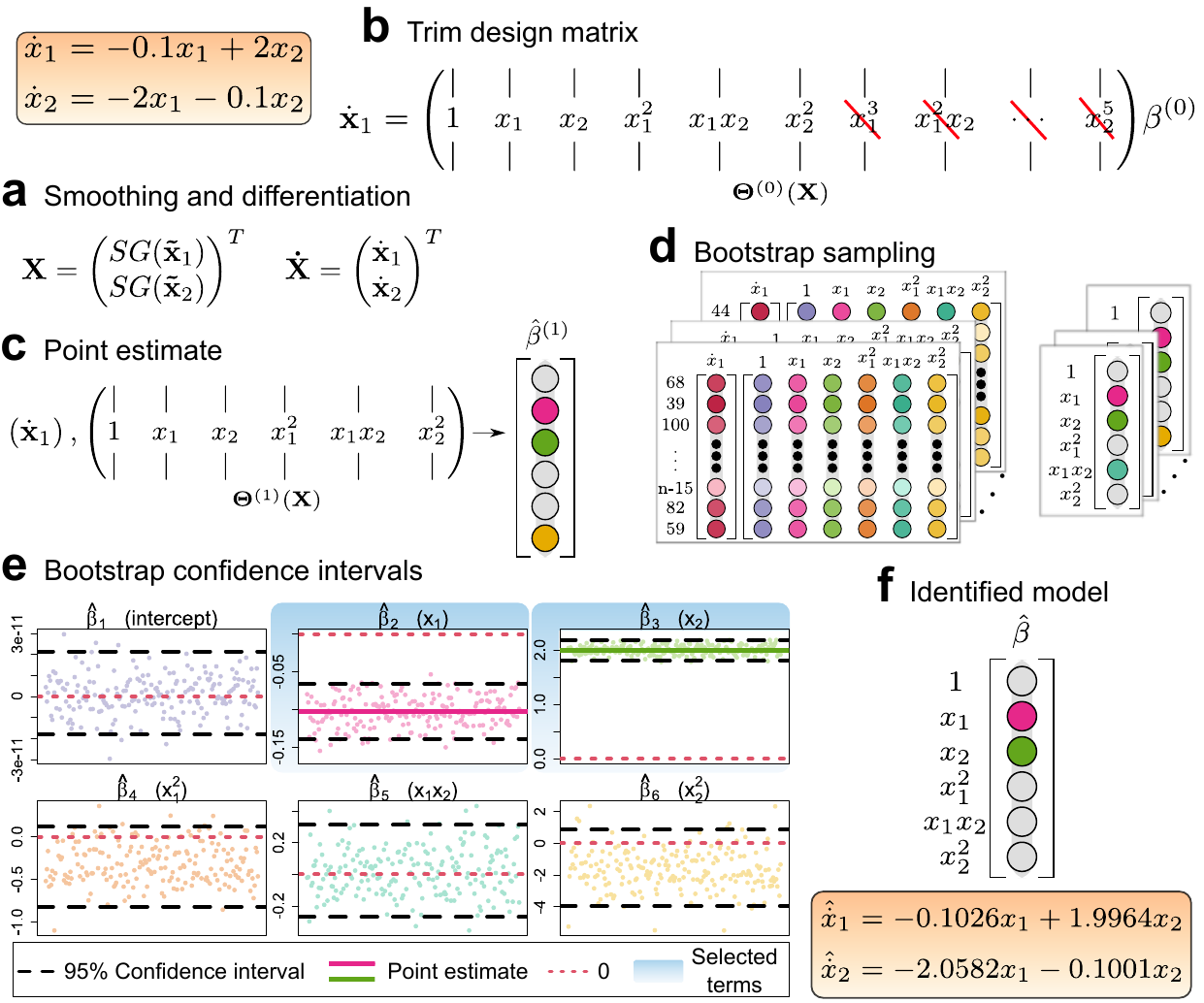}
\caption{\textbf{Automatic regression for governing equations (ARGOS).}
This example illustrates the process of identifying the $\dot{x}_1$ equation of a two-dimensional damped oscillator with linear dynamics.
We first (\textbf{a}) smooth each noisy state vector in $\mathbf{\Tilde{X}}$ and calculate the derivative $\mathbf{\dot{x}}_1$ using the Savitzky-Golay filter.
Next, we (\textbf{b}) construct the design matrix $\mathbf{\Theta}^{(0)}(\mathbf{X})$, containing the observations $x(t)$ and their interaction terms up to monomial degree $d=5$ ---see Eq.~(\ref{theta_matrix}).
Following Eq.~(\ref{argos_system_id}), we perform sparse regression using either the lasso or the adaptive lasso and determine the highest-order monomial degree with a nonzero coefficient in the estimate $\hat{\beta}^{(0)}$ (in this example, we detect terms up to $d=2$).
We then trim the design matrix to include only terms up to this order and (\textbf{c}) perform sparse regression again with the trimmed design matrix and the previously used algorithm (lasso or adaptive lasso), apply OLS on the subset of selected variables, and determine the final $\hat{\beta}^{(1)}$ point estimates.
Finally, we (\textbf{d}) employ bootstrap sampling to obtain 2000 sample estimates and (\textbf{e}) develop 95\% bootstrap confidence intervals to (\textbf{f}) identify the $\hat{\beta}$ by selecting the coefficients whose intervals contain the point estimate but do not include zero.
}
\label{Fig:sparse_regression_schematic}
\end{figure*}

Our approach, ARGOS, aims to automatically identify interpretable models that describe the dynamics of a system by integrating machine learning with statistical inference.
As illustrated in Fig.~\ref{Fig:sparse_regression_schematic}, our algorithm comprises several key phases to solve the system in Eq.~\eqref{derivative_system_id_equation_with_noise}.
These include data smoothing and numerical approximation of derivatives, as well as the use of bootstrap sampling with sparse regression to develop confidence intervals for variable selection.

In the first phase, we employ the Savitzky-Golay filter, which fits a low-degree polynomial to a local window of data points, reducing noise in the state matrix and approximating the derivative numerically~\cite{savitzkySmoothingDifferentiationData1964}.
To optimize the filter, we set polynomial order $o=4$ and construct a grid of window lengths $l$~\cite{h.pressNumericalRecipesArt2007}.
For each column of the noisy state matrix $\mathbf{\Tilde{X}}$, we find the optimal $l^\ast$ that minimizes the mean squared error between each noisy signal $\mathbf{\Tilde{x}}_j$ and its smoothed counterpart $\mathbf{x}_j$ (Supplementary Algorithm~\ref{al:SG}).

Following smoothing and differentiation, we construct the design matrix $\mathbf{\Theta}(\mathbf{X})$ with monomials up to the $d$-th degree and extract the columns of $\mathbf{\dot{X}}$ and $\mathbf{B}$ from Eq.~\eqref{derivative_system_id_equation_with_noise} to identify the governing equations of each component of the system:
\begin{equation}
\label{argos_system_id}
    \mathbf{\dot{x}}_j = \mathbf{\Theta}(\mathbf{X}){\beta_j} + \epsilon_j, \qquad j = 1,\dotsc,m.
\end{equation}
We then apply either the least absolute shrinkage and selection operator (lasso)~\cite{tibshiraniRegressionShrinkageSelection1996} or the adaptive lasso~\cite{zouAdaptiveLassoIts2006} during the model selection process (Supplementary Algorithm~\ref{al:ARGOS}).
Both algorithms add the $\ell_1$ penalty to the ordinary least squares regression (OLS) estimate, shrinking coefficients to zero.
This allows for the selection of the nonzero terms for parameter and model inference.

After identifying an initial sparse regression estimate of $\beta_j$ in Eq.~\eqref{argos_system_id}, we trim the design matrix to include only monomial terms up to the highest-order variable with a nonzero coefficient in the estimate.
Using the updated design matrix, we reapply the sparse regression algorithm and employ a grid of thresholds to develop a subset of models, with each model containing only coefficients whose absolute values exceed their respective thresholds.
Next, we perform OLS on the selected variables of each subset to calculate unbiased coefficients and determine the point estimates from the regression model with the minimum Bayesian information criterion (BIC)~\cite{schwarzEstimatingDimensionModel1978}.
As a final step, we bootstrap this sparse regression process with the trimmed design matrix to obtain 2000 sample estimates~\cite{efronIntroductionBootstrap1993}.
We then construct 95\% bootstrap confidence intervals using these sample estimates and identify a final model consisting of variables whose confidence intervals do not include zero and whose point estimates lie within their respective intervals.

\subsection{Assessing ARGOS systematically}
\label{subsect:benchmarking_models}
\begin{figure}[!htp]
\centering
\includegraphics[width=\textwidth]{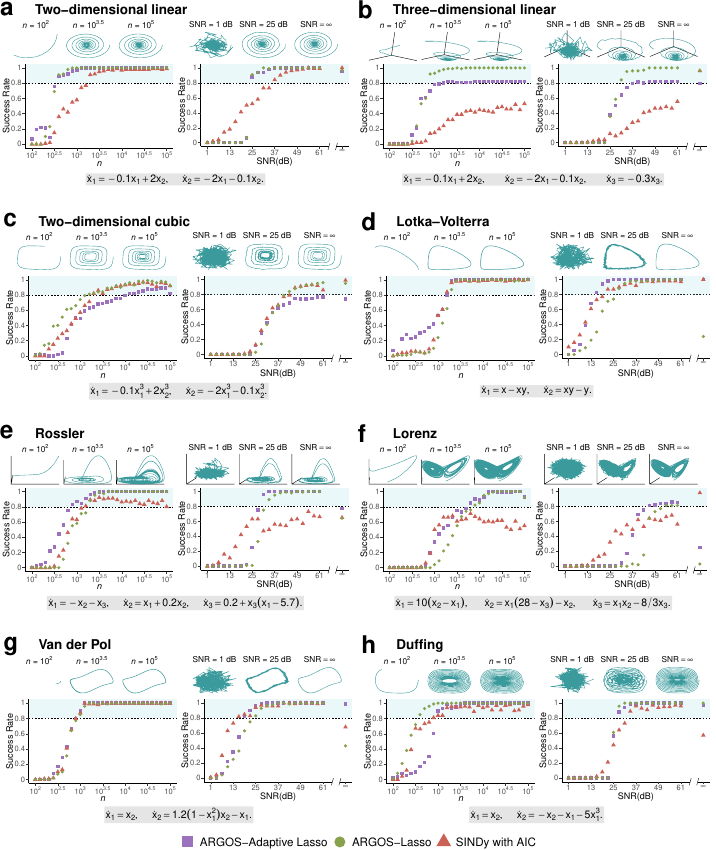}
\caption{\textbf{Success rate of ARGOS versus SINDy with AIC for linear and nonlinear systems.}
We generate 100 random initial conditions and examine the success rate of ARGOS and SINDy with AIC in discovering the correct terms of the governing equations from each system at each value of $n$ and SNR. 
(\textbf{a})-(\textbf{b}) Linear systems.
First-order nonlinear systems in two (\textbf{c})-(\textbf{d}) and three (\textbf{e})-(\textbf{f}) dimensions.
(\textbf{g})-(\textbf{h}) Second-order nonlinear systems. 
We increase the time-series length $n$ while holding $\text{SNR}=49\ \text{dB}$ (left panels) and fix $n=5000$ when increasing the SNR (right panels).
Shaded regions represent model discovery above 80\%.
}
\label{fig:all_systems_success_rate}
\end{figure}

To evaluate the effectiveness of our approach, we used 100 random initial conditions to expand well-known ODEs and generated data sets with varying time series lengths $n$ and signal-to-noise ratios (SNR) (see~\nameref{sect:methods}).
We then introduced a success rate metric, defined as the proportion of instances where an algorithm identified the correct terms of the governing equations from a given dynamical system.
This metric allowed us to quantitatively measure the performance of an algorithm across different dynamical systems, as well as different SNR and $n$ values (see Supplementary Tables~\ref{tab:min_n_table}~and~\ref{tab:min_snr_table}).
Figure~\ref{fig:all_systems_success_rate} highlights success rates exceeding 80\%, demonstrating that our method consistently outperformed SINDy with AIC in identifying the underlying system of the data.
We accurately represented linear systems with less than 800 data points and medium SNRs, underscoring the method's ability to handle straightforward dynamics.
Notably, even with only moderately-sized data sets or medium SNRs, we successfully identified three out of five of the two-dimensional ODEs using the lasso with ARGOS, showcasing the effectiveness of integrating classic statistical learning algorithms within our framework.
The adaptive lasso was able to identify the non-linear ODEs in three dimensions with higher accuracy than the other algorithms tested.
These results suggest that the adaptive lasso is suitable for identifying non-linear ODEs in higher dimensional systems.

\begin{figure*}[!htp]
\centering
\begin{minipage}[c]{0.495\linewidth}
\includegraphics[width=1\linewidth]{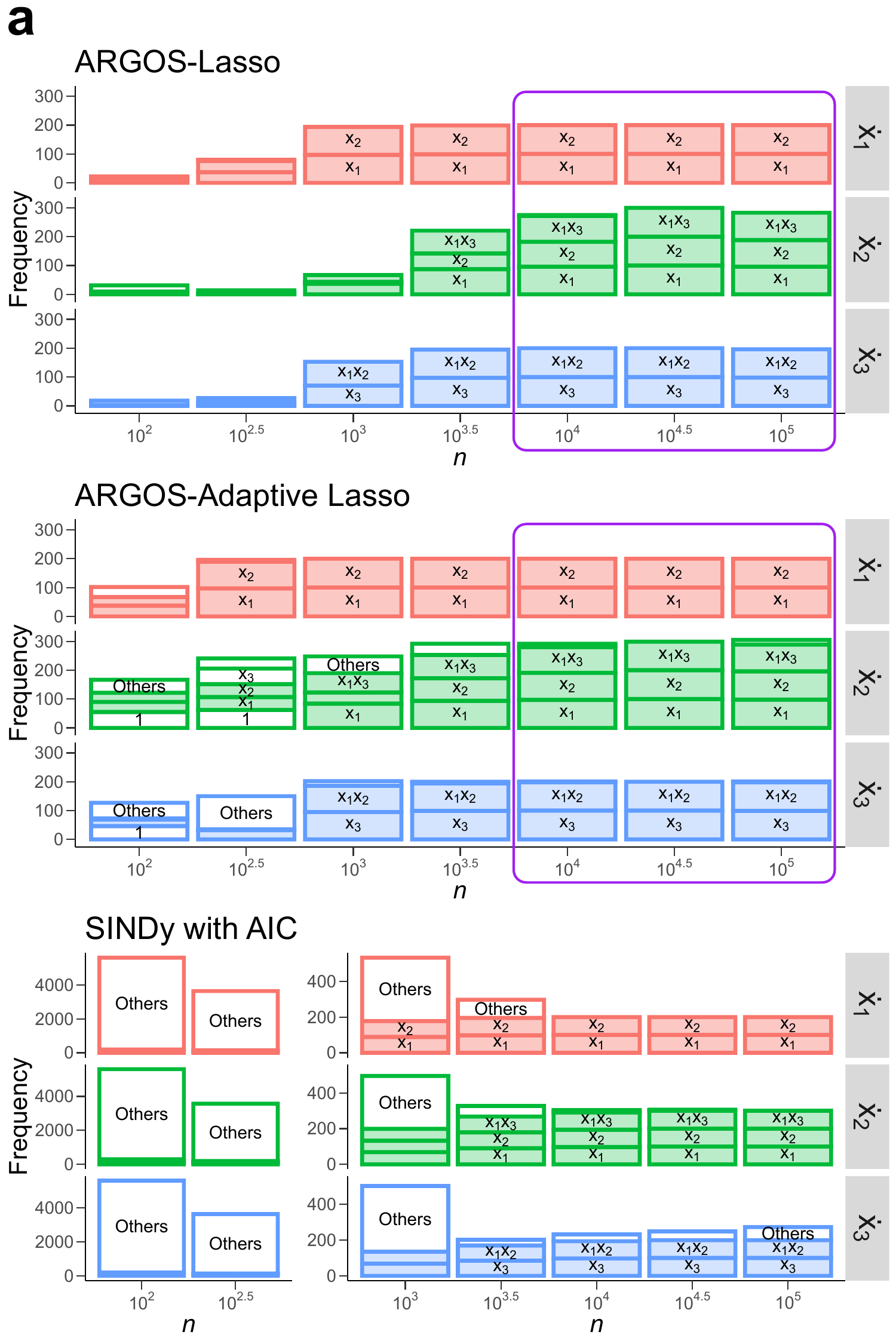}
\end{minipage}
\begin{minipage}[c]{0.495\linewidth}
\includegraphics[width=1\linewidth]{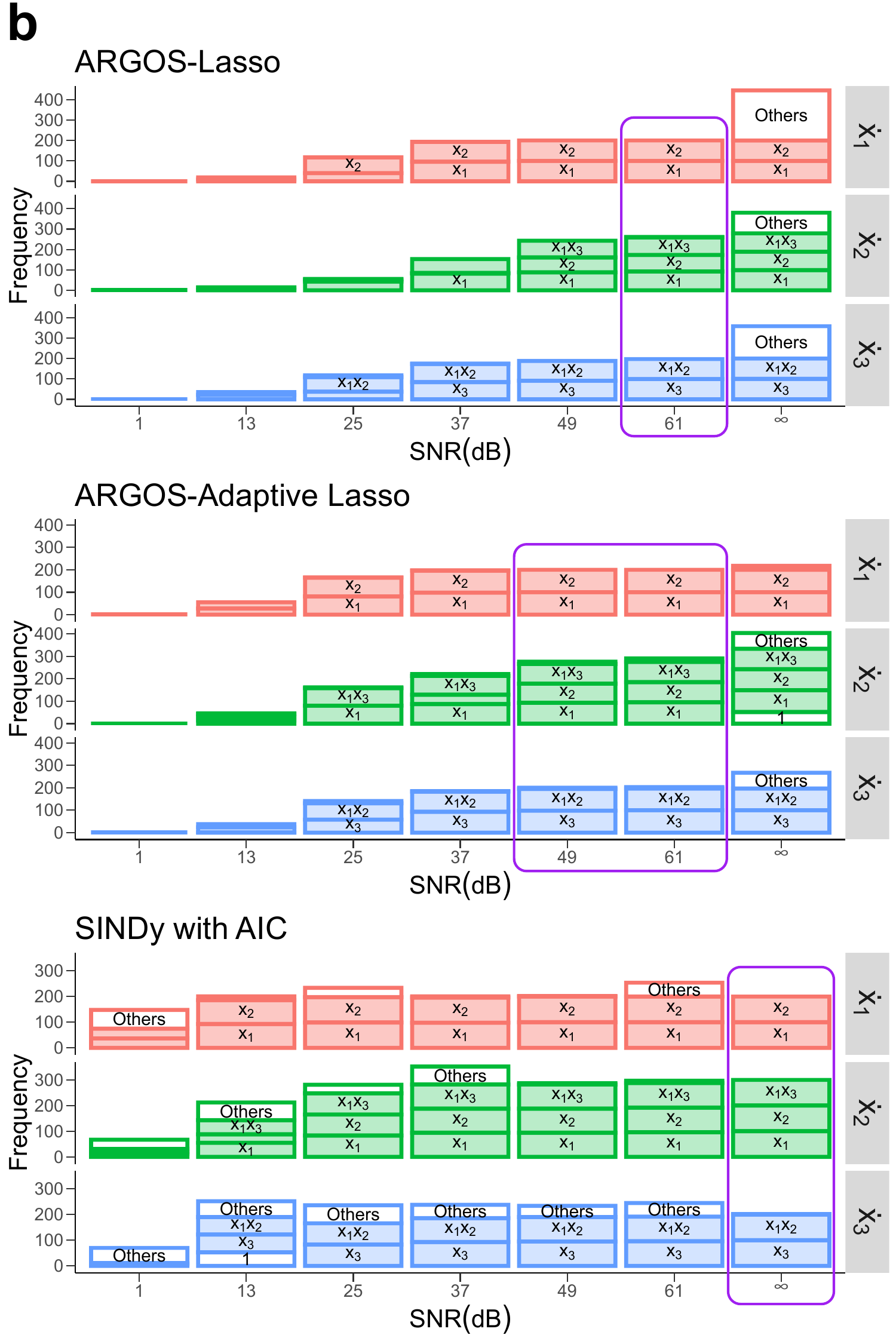}
\end{minipage}
\includegraphics[width=0.4\textwidth]{legend_3d.pdf}
\caption{\textbf{Frequency of identified variables for the Lorenz system across algorithms.}
Colors correspond to each governing equation; filled boxes indicate correctly identified variables, while white boxes denote erroneous terms.
Panels show the frequency of identified variables for data sets with (\textbf{a}) increasing $n$ (SNR = 49 dB), and (\textbf{b}) SNR ($n=5000$).
Purple-bordered regions demarcate model discovery above 80\%.
}
\label{fig:lorenz_stacked}
\end{figure*}

\begin{figure}[htp]
\centering
\includegraphics[width=0.48\textwidth]{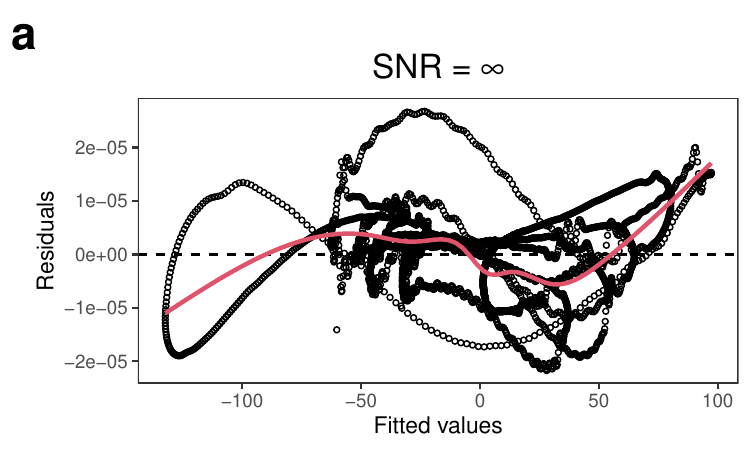}
\includegraphics[width=0.48\textwidth]{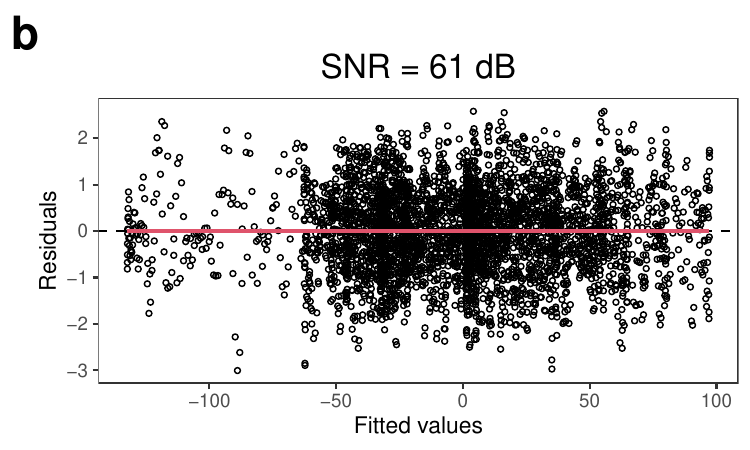}
\caption{\textbf{Residuals vs fitted diagnostics for the ARGOS-Lasso identified model of the Lorenz $\dot{x}_1$ equation}.
Comparison of residuals for the prediction models identified using the lasso with ARGOS for the Lorenz system's $\dot{x}_1$ equation when data are (\textbf{a}) noiseless and (\textbf{b}) contaminated by $\text{SNR}= 61\ \text{dB}$.
}
\label{fig:autocorrelation}
\end{figure}

The systematic analysis, presented in Fig.~\ref{fig:all_systems_success_rate}, emphasized the efficacy of our approach as $n$ and SNR increased.
The importance of data quality and quantity is further supported by Fig.~\ref{fig:lorenz_stacked}, which illustrates the frequency at which our approach identified each term in the design matrix across different values of $n$ and SNR.
The boxes in the figure delineate regions where each algorithm achieved model discovery above 80\% for the Lorenz system, providing insights into the selected terms contributing to the success and failure of each method across different settings.
When faced with limited observations and low signal quality, our approach identified overly sparse models that failed to represent the governing dynamics accurately, while SINDy with AIC selected erroneous terms, struggling to obtain a parsimonious representation of the underlying equations.
Figure~\ref{fig:lorenz_stacked} also illustrates the decline in our method's performance for deterministic systems, as it identified several ancillary terms for the Lorenz dynamics when $\text{SNR}=\infty$ (additional details for each system are provided in the \nameref{Supplementary_Information}).
The decrease in identification accuracy stemmed from the identified model's violation of the homoscedasticity assumption in linear regression, which occurs when residuals exhibit non-constant variance.
Figure~\ref{fig:autocorrelation} demonstrates that our method did not satisfy this assumption when identifying the $\dot{x}_1$ equation of the Lorenz system.
Consequently, our approach selected additional terms to balance the variance among the model's residuals while sacrificing correct system discovery.
As the noise in the system slightly increased, however, homoscedasticity in the residuals became more pronounced, enabling our approach to distinguish the equation's true underlying structure.
Thus, our method proved more practical in accurately identifying the correct terms of the governing equations when data contained low levels of noise in the signal, which is often the case in many real-world applications, as opposed to when dealing with noiseless systems.

Our method outperformed SINDy with AIC in identifying a range of ODEs, especially three-dimensional systems.
One potential explanation for the lesser performance of SINDy with AIC is that multicollinearity in the design matrix often causes OLS to produce unstable coefficients.
Due to the sensitivity of the estimated coefficients, small changes in the data can lead to fluctuations in their magnitude, making it difficult for the sparsity-promoting parameter to determine the correct model.
As a result, the initial phase of the hard-thresholding procedure of SINDy with AIC inadvertently removed the true dynamic terms of the underlying system.
Therefore, this model selection approach will likely face persistent challenges when discovering higher-dimensional systems that contain additional multicollinearity in the design matrix.

\begin{figure}[htp]
\centering
\includegraphics[width=0.48\textwidth]{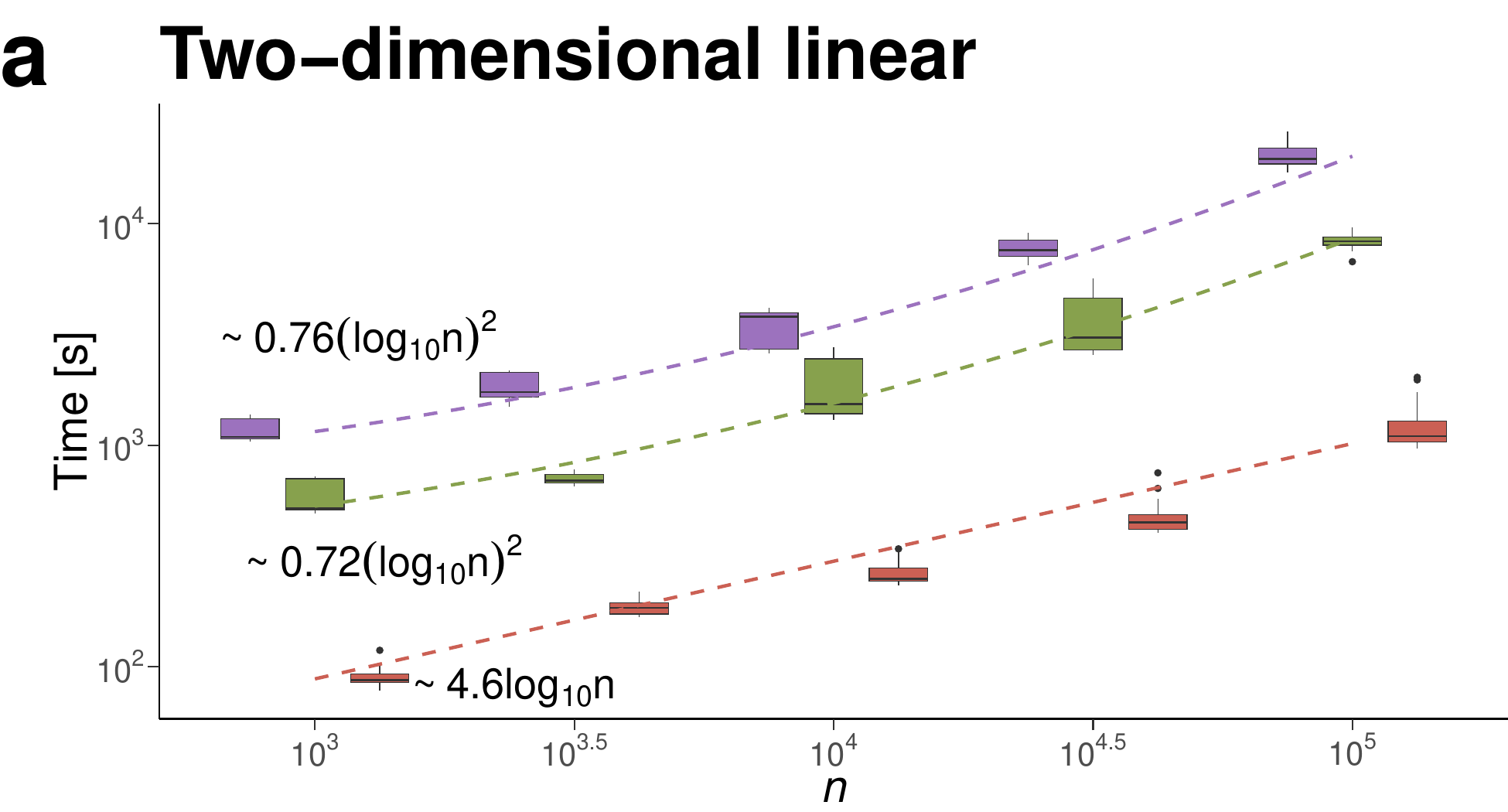}
\includegraphics[width=0.48\textwidth]{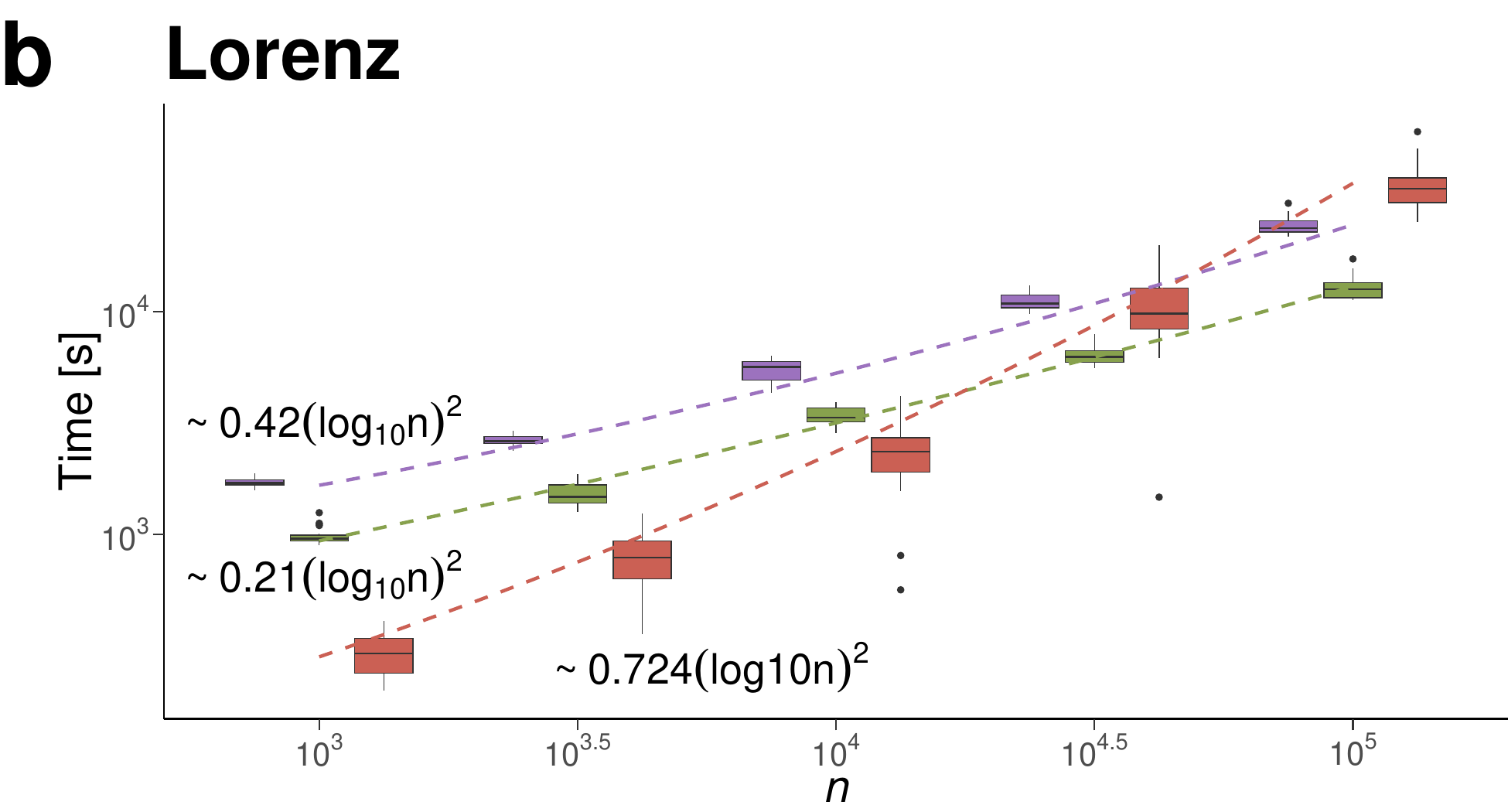}\\
\includegraphics[width=0.48\textwidth]{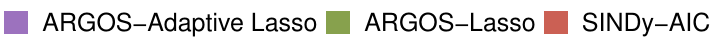}
\caption{\textbf{Time-complexity (seconds) between ARGOS and SINDy with AIC}.
Boxplots depict the computational time required for model discovery over 30 instances for (\textbf{a}) a two-dimensional damped oscillator with linear dynamics and (\textbf{b}) the Lorenz system.
The black bar within each box represents the median computational time.
Whiskers extending from each box show 1.5 times the interquartile range.
Data points beyond the end of the whiskers are outlying points.
Equations accompanying the dashed lines indicate the fitted mean computational time for each algorithm at various values of $n$.
}
\label{fig:runtimes}
\end{figure}

Figure~\ref{fig:runtimes} shows the computational time, measured in seconds, required for our approach and SINDy with AIC to perform model discovery.
While our method demanded greater computational effort for the two-dimensional linear system than SINDy with AIC, it demonstrated better efficiency in identifying the Lorenz dynamics as $n$ increased.
The decrease in efficiency of SINDy with AIC can be attributed to its model selection process, which involves enumerating all potential prediction models---a procedure that becomes progressively more expensive with data in higher dimensions~\cite{manganModelSelectionDynamical2017}.
In contrast, our approach displayed a similar rate of increase in computational complexity as the time series expanded for both systems, suggesting that our method was less affected by the growing data dimensionality than SINDy with AIC.
Thus, our method offers a more efficient alternative for identifying three-dimensional systems with increasing time series lengths.

\section{Discussion}
\label{sect:discussion}
We have demonstrated an automatic method, ARGOS, for extracting dynamical systems from scarce and noisy data without prior knowledge of the governing equations.
Our approach combines the Savitzky-Golay filter for signal denoising and differentiation with sparse regression and bootstrap sampling for confidence interval estimation, effectively addressing the inverse problem of inferring underlying dynamics from observational data through reliable variable selection.
By examining diverse trajectories, we showcased the capabilities of our algorithm in automating the discovery of mathematical models from data, consistently outperforming the established SINDy with AIC, especially when identifying systems in three dimensions.

While we have shown promising results with our approach, it is important to note several potential limitations.
First, although our method effectively automates model discovery, it can only correctly represent the true governing equations if the active terms are present in the design matrix, a constraint inherent in regression-based identification procedures.
Building on this point, we stress the importance of data quantity and quality as identification accuracy improved with sufficient observations and moderate to high signal-to-noise ratios.
We also found that our method performs better when data contains low levels of noise, as opposed to noiseless systems. The linear regression assumption of homoscedasticity is violated under noiseless conditions, and the method identifies spurious terms to develop a more constant variance among the residuals. 
However, this issue can be mitigated in the presence of a small amount of noise in the data, leading to a more constant variance in the residuals of the true model and enabling more accurate identification.
Lastly, as the number of observations and data dimensionality increase, bootstrap sampling becomes computationally demanding, which can significantly prolong the model selection process and limit our algorithm's applicability in real-time.
Nonetheless, obtaining confidence intervals through bootstrap sampling serves as a reliable approach for our method, allowing us to eliminate superfluous terms and select the ones that best represent the underlying equations, ultimately leading to more accurate predictions of the system's dynamics.

In this information-rich era, data-driven methods for uncovering governing equations are increasingly crucial in scientific research.
By developing automated processes, researchers can develop concise models that accurately represent the underlying dynamics in their data, accelerating advancements across various disciplines in science.
Our study endorses an inference-based approach that combines statistical learning and model assessment methods, emphasizing the importance of thorough model evaluation for building confidence in discovering governing equations from data.
As data-driven techniques advance, we look forward to further developments in automatic system identification that will continue contributing to the search for the elusive laws governing many intricate systems.

\section{Methods}
\label{sect:methods}
\subsection{The lasso and adaptive lasso for variable selection}
\label{subsect:sparse_regression}
For the $j$th column of $\mathbf{\dot{X}}$ and $\mathbf{B}$ in Eq.~\eqref{derivative_system_id_equation_with_noise}, we implement sparse regression by adding weighted $\ell_q$ penalties to the OLS regression estimate~\cite{tibshiraniElementsStatisticalLearning2009}
\begin{equation}
\label{optimisation}
\argmin_{\beta} \left\{\sum_{i=1}^n \left (\dot{x}_i-\beta_0 - \sum_{k=1}^p\theta(\mathbf{X})_{i,k}\beta_{k}\right )^2 + \lambda\sum_{k=1}^p w_{k}\abs{\beta_{k}}^q \right\}.
\end{equation}
When all weights $w_{k} = 1$ for $k=1,\dots, p$, Eq.~\eqref{optimisation} represents the lasso for $q=1$ and ridge regression for $q=2$~\cite{tibshiraniRegressionShrinkageSelection1996, hoerlRidgeRegressionBiased1970}.
Furthermore, the adaptive lasso is derived from the lasso by incorporating pilot estimates $\Tilde{\beta}$ and setting $w_{k} = 1/\abs{\tilde{\beta}_{k}}^{\nu}$~\cite{zouAdaptiveLassoIts2006}.
The weighted penalty in the adaptive lasso can be interpreted as an approximation of the $\ell_p$ penalties with $p=1-\nu$~\cite{hastieStatisticalLearningSparsity2015}.
Therefore, fixing $\nu=1$ allows us to achieve a soft-threshold approximation to the $\ell_0$ penalty, providing an alternative to the hard-thresholding in the SINDy algorithm, which requires a choice of the cut-off hyperparameter~\cite{bruntonDiscoveringGoverningEquations2016}.

As $\lambda$ increases in Eq.~\eqref{optimisation}, ridge regression, the lasso, and the adaptive lasso shrink the coefficients toward zero.
However, of these three methods, the lasso and the adaptive lasso perform variable selection by reducing small coefficients to exactly zero~\cite{tibshiraniRegressionShrinkageSelection1996}.
We use \texttt{glmnet} to solve Eq.~\eqref{optimisation} by producing a default $\lambda$ grid and applying 10-fold cross-validation to determine the optimal initial tuning parameter $\lambda_0^\ast$~\cite{friedmanRegularizationPathsGeneralized2010}.
We then refine the grid around $\lambda_{0}^\ast$ with 100 points spanning $[\lambda_{0}^\ast / 10, 1.1 \cdot \lambda_{0}^\ast]$ and impose this updated grid on \texttt{glmnet} to solve Eq.~\eqref{optimisation} again, identifying the optimal $\lambda^\ast$ that best predicts $\mathbf{\dot{x}}_j$.

The lasso is effective when only a few coordinates of the coefficients $\beta$ are nonzero.
However, like OLS, the lasso provides unstable estimates when predictors are collinear, whereas ridge regression produces more stable solutions when multicollinearity exists in the data~\cite{tibshiraniElementsStatisticalLearning2009}.
Therefore, we apply ridge regression to the data in the first stage of the adaptive lasso to obtain stable pilot estimates $\Tilde{\beta}$ and reduce the effects of multicollinearity~\cite{zouAdaptiveLassoIts2006}.

The second stage of the adaptive lasso then uses the $\Tilde{\beta}$ pilot estimates to calculate the weights vector $w$, enabling variable selection by solving the problem in Eq.~\eqref{optimisation}.
Here, we calculate the weights vector $w$ using pilot estimates $\Tilde{\beta}$ corresponding to the optimal $\lambda^*_{\text{ridge}}$ ridge regression model before identifying a separate tuning parameter $\lambda^\ast_{\text{adaptive lasso}}$.
In doing so, we make Eq.~\eqref{optimisation} less computationally expensive since we optimize twice on a single parameter rather than simultaneously optimizing over $\lambda^\ast_{\text{ridge}}$ and $\lambda^\ast_{\text{adaptive lasso}}$~\cite{buhlmannStatisticsHighDimensionalData2011}.

The adaptive lasso often yields a sparser solution than the lasso since applying individual weights to each variable places a stronger penalty on smaller coefficients, reducing more of them to zero.
Here, small $\Tilde{\beta}$ coefficients from the first stage of the adaptive lasso lead to a larger penalty in the second.
Larger penalty terms in the second stage of the adaptive lasso result in more coefficients being set to zero than the standard lasso method.
Furthermore, a smaller penalty term enables the adaptive lasso to uncover the true coefficients and reduce bias in the solution~\cite{buhlmannStatisticsHighDimensionalData2011}.

The adaptive lasso, valuable for system identification, obtains the oracle property when the $\Tilde{\beta}$ pilot estimates converge in probability to the true value of $\beta$ at a rate of $1/\sqrt{n}$ ($\sqrt{n}$-consistency).
As $n$ increases, the algorithm will select the true nonzero variables and estimate their coefficients as if using maximum likelihood estimation~\cite{zouAdaptiveLassoIts2006}.

\subsection{Algorithm implementation}
\label{subsect:algo_implementation}
When applying ARGOS for model selection, we use $\eta=10^{-8},10^{-7},\dotsc,10^{1}$ for thresholding the sparse regression coefficients before performing OLS on each subset ${\mathcal{K}_i = \{k:\abs{\hat{\beta}_{k}} \geq \eta_i\}},\ i =1,\dotsc,\mathbf{card}(\eta)$ of selected variables, determining an unbiased estimate for $\beta$~\cite{tibshiraniElementsStatisticalLearning2009}.
We then calculate the BIC for each $\eta$ regression model and select the model with the minimum value, further promoting sparsity in the identification process~\cite{zouDegreesFreedomLasso2007,schwarzEstimatingDimensionModel1978}.

The number of bootstrap sample estimates $B$ must be large enough to develop confidence intervals for variable selection~\cite{efronIntroductionBootstrap1993}.
Therefore, we collect $B=2000$ bootstrap sample estimates and sort them by ${\hat{\beta}_{k}^{\text{OLS}\{1\}}\leq \hat{\beta}_{k}^{\text{OLS}\{2\}}\leq \dots\leq \hat{\beta}_{k}^{\text{OLS}\{B\}}}$.
We then use the 100(1 - $\alpha$)$\%$ confidence level, where $\alpha=0.05$, to calculate the integer part of $B\alpha/2$ and develop estimates of the lower and upper bounds: $CI_\text{lo} = [B\alpha/2]$ and $CI_\text{up} = B-CI_\text{lo}+1$.
Finally, we implement these calculations to develop confidence intervals $\left[\hat{\beta}_{k}^{\text{OLS}\{CI_\text{lo}\}}, \hat{\beta}_{k}^{\text{OLS}\{CI_\text{up}\}} \right]$ from our sample distribution~\cite{zoubirBootstrapTechniquesSignal2004}.

We apply our implementation of the Savitzky-Golay filtering method to perform the entire system identification process with SINDy with AIC automatically.

\subsection{Building the data sets and tests}
\label{subsect:problem_setup}
We conducted two sets of numerical experiments to assess the impact of data quality and quantity on the performance of the  algorithms.
To evaluate the algorithms' performance with limited data, we first kept the signal-to-noise ratio constant ($\text{SNR} = 49$) and increased the number of observations $n$ for each ODE system.
We generated 100 random initial conditions and used temporal grids starting with $t_\text{initial} = 0$ and a varying $t_\text{final}$ between $1$ ($n=10^2$) and $1000$ ($n=10^5$) with a time step $\Delta t = 0.01$.
For the Lorenz equations, we used $\Delta t = 0.001$, resulting in $t_\text{final}$ values ranging from $0.1$ ($n=10^2$) and $100$ ($n=10^5$)~\cite{bruntonDiscoveringGoverningEquations2016}.

To examine the algorithms' performance under noisy conditions, we varied the SNR in the data by corrupting the state matrix with a zero-mean Gaussian noise matrix $\mathbf{Z}\sim\mathcal{N}(0,\sigma_{\mathbf{Z}}^2)$.
In this setting, we determined the standard deviation $\sigma_{\mathbf{z}_j}$ of each column of $\mathbf{Z}$ as
\begin{equation}
\label{sigma_noise}
    \sigma_{\mathbf{z}_j} = \sigma_{\mathbf{x}_j}\cdot 10^{-\frac{\text{SNR}}{20}}, \qquad j = 1,\dotsc,m,
\end{equation}
and develop the noise corrupted $\tilde{\mathbf{X}}$ as~\cite{lyonsUnderstandingDigitalSignal2011}
\begin{equation}
\label{signal_with_noise}
    \tilde{\mathbf{X}} = \mathbf{X} + \mathbf{Z}.
\end{equation}
Keeping $n$ constant, we again used 100 random initial conditions and generated $\tilde{\mathbf{X}}$ matrices increasing in noise levels such that $\text{SNR} = 1,4,\dotsc,61\ \text{dB}$ with $\Delta$SNR = 3 dB, including a noiseless system ($\text{SNR} = \infty$).

For both the tests with varying $n$ and SNR, we constructed the design matrix $\mathbf{\Theta}^{(0)}(\mathbf{X})$ with monomial functions up to $d=5$ of the smoothed columns of $\mathbf{X}$~\cite{bruntonDiscoveringGoverningEquations2016}.
We then performed system identification with each data set and calculated the success rate of each algorithm as the probability of extracting the correct terms of the governing equations.
Additionally, we analyzed the most frequently selected variables of each method.

Finally, we measured the computational time, in seconds, for running our method and SINDy with AIC by performing model discovery on 30 instances of the two-dimensional linear system and the Lorenz system for time series lengths $n=10^2,10^{2.5},\dots,10^5$, using one CPU core with a single thread.

\subsection{Code availability}
\label{subsect:code_avail}
\sloppy All code used in this study is available at \url{http://github.com/kevinegan31/ARGOS}.
\subsection{Data availability}
\label{subsect:data}
\sloppy All data generated for this study can be generated with the code available at \url{http://github.com/kevinegan31/ARGOS}.
We provide a snapshot of the data on GitHub.


\clearpage
\newcommand{\beginsupplement}{%
        \setcounter{section}{0}
        \renewcommand{\thesection}{S\arabic{section}}
        \renewcommand{\theHsection}{S\arabic{section}}
        \newcounter{SIsec}
        \renewcommand{\theSIsec}{S\arabic{SIsec}}
        \setcounter{table}{0}
        \renewcommand{\thetable}{S\arabic{table}}
        \renewcommand{\theHtable}{S\arabic{table}}
        \newcounter{SItab}
        \renewcommand{\theSItab}{S\arabic{SItab}}
        \setcounter{figure}{0}
        \renewcommand{\thefigure}{S\arabic{figure}}
        \renewcommand{\theHfigure}{S\arabic{figure}}
        \newcounter{SIfig}
        \renewcommand{\theSIfig}{S\arabic{SIfig}}
        \setcounter{equation}{0}
        \renewcommand{\theequation}{S\arabic{equation}}
        \renewcommand{\theHequation}{S\arabic{equation}}
        \newcounter{SIeq}
        \renewcommand{\theSIeq}{S\arabic{SIeq}}
     }

\beginsupplement
\vskip6pt

\enlargethispage{20pt}

\section{Supplementary Information}\label{Supplementary_Information}
\begin{table}[!htp]
    \centering
    \caption{Minimum number of observations ($n$) needed for each method to obtain 80\% accuracy in identifying governing equations of dynamical systems.
    Top-performing algorithms are in red, and three-dimensional systems have a shaded background.}
    \begin{tabularx}{0.63\linewidth}{l|l|l}
        \toprule
        System & Algorithm & $n$ \\
        \midrule
        \multirow{3}{*}{Two-dimensional linear} & \textcolor{myred}{ARGOS-Lasso} & $10^{2.6}\ (399)$ \\
        & \textcolor{myred}{ARGOS-Adaptive Lasso} & $10^{2.6}\ (399)$ \\
        & SINDy with AIC & $10^{3.3}\ (1996)$ \\
        \midrule
        \cellcolor{highlight} & \cellcolor{highlight}\textcolor{myred}{ARGOS-Lasso} & \cellcolor{highlight}$10^{2.9}\ (795)$ \\
        \cellcolor{highlight} & \cellcolor{highlight}ARGOS-Adaptive Lasso & \cellcolor{highlight}$10^{3.2}\ (1585)$ \\
        \cellcolor{highlight}\multirow{-3}{*}{Three-dimensional linear} & \cellcolor{highlight}SINDy with AIC & \cellcolor{highlight}NA \\
        \midrule
        \multirow{3}{*}{Two-dimensional cubic} & \textcolor{myred}{ARGOS-Lasso} & $10^{3.2}\ (1585)$ \\
        & SINDy with AIC & $10^{3.3}\ (1996)$ \\
        & ARGOS-Adaptive Lasso & $10^{4.1}\ (12590)$ \\
        \midrule
        \multirow{3}{*}{Lotka-Volterra} & \textcolor{myred}{ARGOS-Adaptive Lasso} & $10^{3.2}\ (1585)$ \\
        & \textcolor{myred}{SINDy with AIC} & $10^{3.2}\ (1585)$ \\
        & ARGOS-Lasso & $10^{3.3}\ (1996)$ \\
        \midrule
        \cellcolor{highlight} & \cellcolor{highlight}\textcolor{myred}{ARGOS-Adaptive Lasso} & \cellcolor{highlight}$10^{2.9}\ (795)$ \\
        \cellcolor{highlight} & \cellcolor{highlight}ARGOS-Lasso & \cellcolor{highlight}$10^{3.2}\ (1585)$ \\
        \cellcolor{highlight}\multirow{-3}{*}{Rossler} & \cellcolor{highlight}SINDy with AIC & \cellcolor{highlight}$10^{3.2}\ (1585)$ \\
        \midrule
        \cellcolor{highlight} & \cellcolor{highlight}\textcolor{myred}{ARGOS-Adaptive Lasso} & \cellcolor{highlight}$10^{3.8}\ (6310)$ \\
        \cellcolor{highlight} & \cellcolor{highlight}ARGOS-Lasso & \cellcolor{highlight}$10^{3.9}\ (7944)$ \\
        \cellcolor{highlight}\multirow{-3}{*}{Lorenz} & \cellcolor{highlight}SINDy with AIC & \cellcolor{highlight}NA \\
        \midrule
        \multirow{3}{*}{Van der Pol} & \textcolor{myred}{ARGOS-Adaptive Lasso} & $10^{2.9}\ (795)$ \\
        & \textcolor{myred}{SINDy with AIC} & $10^{2.9}\ (795)$ \\
        & ARGOS-Lasso & $10^{3.0}\ (1000)$ \\
        \midrule
        \multirow{3}{*}{Duffing} & \textcolor{myred}{ARGOS-Lasso} & $10^{2.6}\ (399)$ \\
        & SINDy with AIC & $10^{2.9}\ (795)$ \\
        & ARGOS-Adaptive Lasso & $10^{3.0}\ (1000)$ \\
        \bottomrule
    \end{tabularx}
\label{tab:min_n_table}
\end{table}

\begin{table}[!htp]
    \centering
    \caption{Maximum signal-to-noise ratio (SNR) tolerated by each method to achieve 80\% accuracy in identifying the governing equations of the dynamical systems.
    Top-performing algorithms are in red, and three-dimensional systems have a shaded background.}
    \begin{tabularx}{0.55\linewidth}{l|l|c}
        \toprule
        System & Algorithm & SNR \\
        \midrule
        \multirow{3}{*} {Two-dimensional linear} & \textcolor{myred}{ARGOS-Lasso} & 25 \\
        & \textcolor{myred}{ARGOS-Adaptive Lasso} & 25 \\
        & SINDy with AIC & 37 \\
        \midrule
        \cellcolor{highlight} & \cellcolor{highlight}\textcolor{myred}{ARGOS-Lasso} & \cellcolor{highlight}31 \\
        \cellcolor{highlight} & \cellcolor{highlight}ARGOS-Adaptive Lasso & \cellcolor{highlight}40 \\
        \cellcolor{highlight}\multirow{-3}{*}{Three-dimensional linear} & \cellcolor{highlight}SINDy with AIC & \cellcolor{highlight}$\infty$ \\
        \midrule
        \multirow{3}{*}{Two-dimensional cubic} & \textcolor{myred}{ARGOS-Lasso} & 43 \\
        & SINDy with AIC & 46 \\
        & ARGOS-Adaptive Lasso & NA \\
        \midrule
        \multirow{3}{*}{Lotka-Volterra} & \textcolor{myred}{ARGOS-Adaptive Lasso} & 16 \\
        & SINDy with AIC & 22 \\
        & ARGOS-Lasso & 28 \\
        \midrule
        \cellcolor{highlight} & \cellcolor{highlight}\textcolor{myred}{ARGOS-Adaptive Lasso} & \cellcolor{highlight}31 \\
        \cellcolor{highlight} & \cellcolor{highlight}ARGOS-Lasso & \cellcolor{highlight}34 \\
        \cellcolor{highlight}\multirow{-3}{*}{Rossler} & \cellcolor{highlight}SINDy with AIC & \cellcolor{highlight}NA \\
        \midrule
        \cellcolor{highlight} & \cellcolor{highlight}\textcolor{myred}{ARGOS-Adaptive Lasso} & \cellcolor{highlight}46 \\
        \cellcolor{highlight} & \cellcolor{highlight}ARGOS-Lasso & \cellcolor{highlight}55 \\
        \cellcolor{highlight}\multirow{-3}{*}{Lorenz} & \cellcolor{highlight}SINDy with AIC & \cellcolor{highlight}$\infty$ \\
        \midrule
        \multirow{3}{*}{Van der Pol} & \textcolor{myred}{SINDy with AIC} & 16 \\
        & ARGOS-Adaptive Lasso & 19 \\
        & ARGOS-Lasso & 25 \\
        \midrule
        \multirow{3}{*}{Duffing} & \textcolor{myred}{ARGOS-Lasso} & 28 \\
        & \textcolor{myred}{ARGOS-Adaptive Lasso} & 28 \\
        & SINDy with AIC & 34 \\
        \bottomrule
    \end{tabularx}
\label{tab:min_snr_table}
\end{table}

\section{Linear systems}
\label{supp_info:linear_systems}
\subsection*{Two-dimensional damped oscillator with linear dynamics}
\label{subsupp_sect:linear2D}
We examined the two-dimensional linear system as~\cite{bruntonDiscoveringGoverningEquations2016}
\begin{equation}
\begin{aligned}
\label{eq:linear2D_system}
    \dot{x}_1 &= -0.1x_1 + 2x_2, \\
    \dot{x}_2 &= -2x_1 - 0.1x_2 .
\end{aligned}
\end{equation}
\begin{figure*}[!htp]
\centering
\begin{minipage}[c]{0.495\linewidth}
\includegraphics[width=1\linewidth]{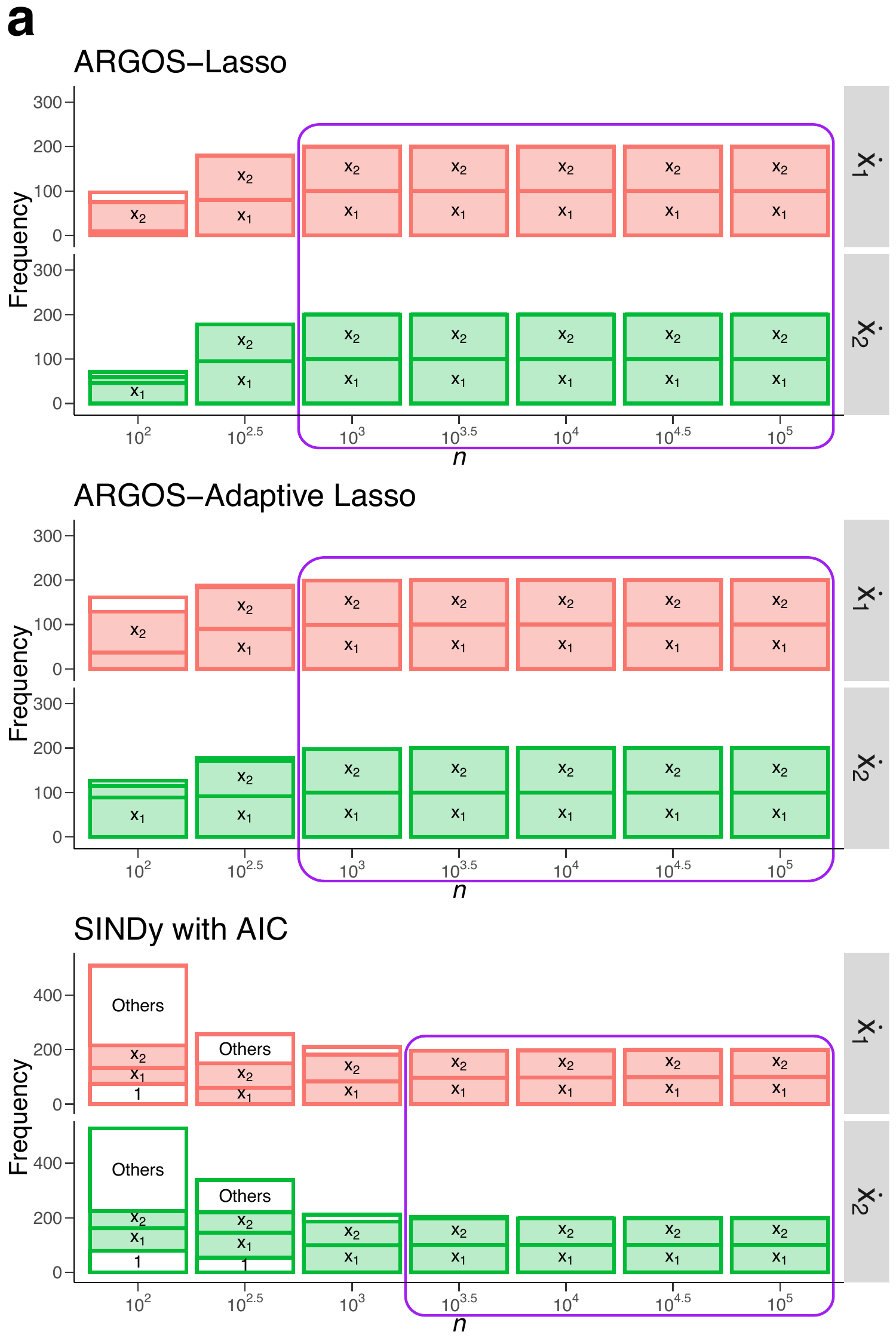}
\end{minipage}
\begin{minipage}[c]{0.495\linewidth}
\includegraphics[width=1\linewidth]{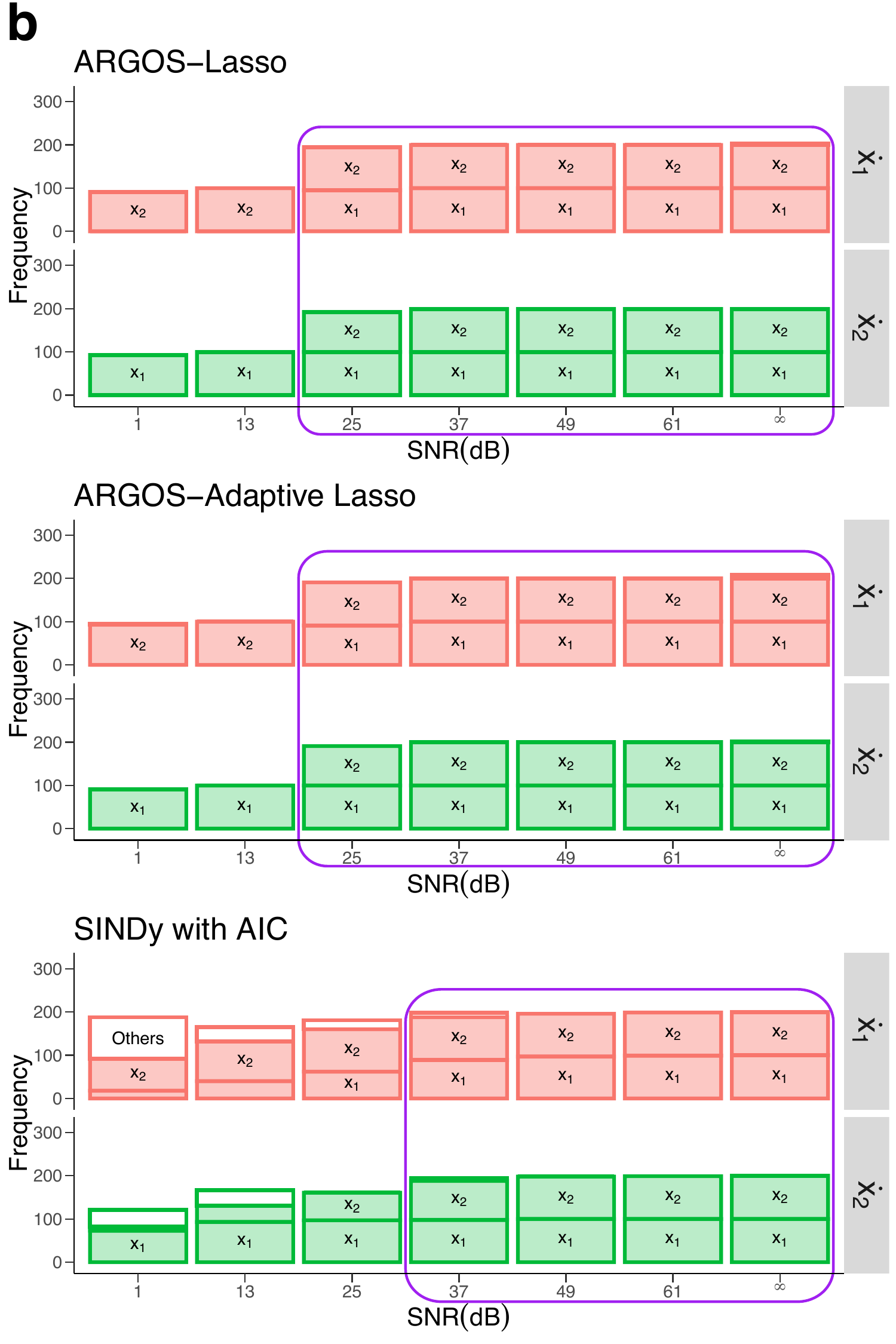}
\end{minipage}
\includegraphics[width=0.3\textwidth]{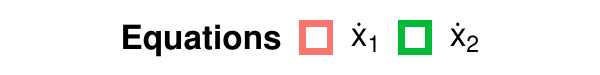}
\caption{\textbf{Frequency of identified variables for the two-dimensional damped oscillator with linear dynamics across algorithms.}
Colors correspond to each governing equation; filled boxes indicate correctly identified variables, while white boxes denote erroneous terms.
Panels show the frequency of identified variables for data sets with (\textbf{a}) increasing $n$ (SNR = 49 dB), and (\textbf{b}) SNR ($n=5000$).
Purple-bordered regions demarcate model discovery above 80\%.
}
\label{fig:linear2d_stacked}
\end{figure*}
For $x_1(t)$ and $x_2(t)$, we generated a random uniform distribution of 100 values between $[10^{-1}, 10^3]$.

In Fig.~\ref{fig:linear2d_stacked}, we observe the performance of our approach and SINDy with AIC in discovering the two-dimensional damped oscillator with linear dynamics.
We found that with less than 300 observations and low SNR, our method identified overly sparse models and struggled to represent the underlying equations of the system accurately.
As the length of the time series $n$ increased and the data became less contaminated with noise, however, the performance of our method improved in extracting the true terms.
Conversely, SINDy with AIC demonstrated a tendency to produce dense models, which contained numerous erroneous variables, particularly with less than 1000 observations and low to medium SNR.

\subsection*{Three-dimensional linear system}
\label{subsupp_sect:linear3D}
We evaluated a three-dimensional system~\cite{bruntonDiscoveringGoverningEquations2016}:
\begin{equation}
\label{eq:linear3D_system}
\begin{aligned}
    \dot{x}_1 &= -0.1x_1 + 2x_2 ,\\
    \dot{x}_2 &= -2x_1 - 0.1x_2 ,\\
    \dot{x}_3 &= -0.3x_3.
\end{aligned}
\end{equation}
\begin{figure*}[!htp]
\centering
\begin{minipage}[c]{0.495\linewidth}
\includegraphics[width=1\linewidth]{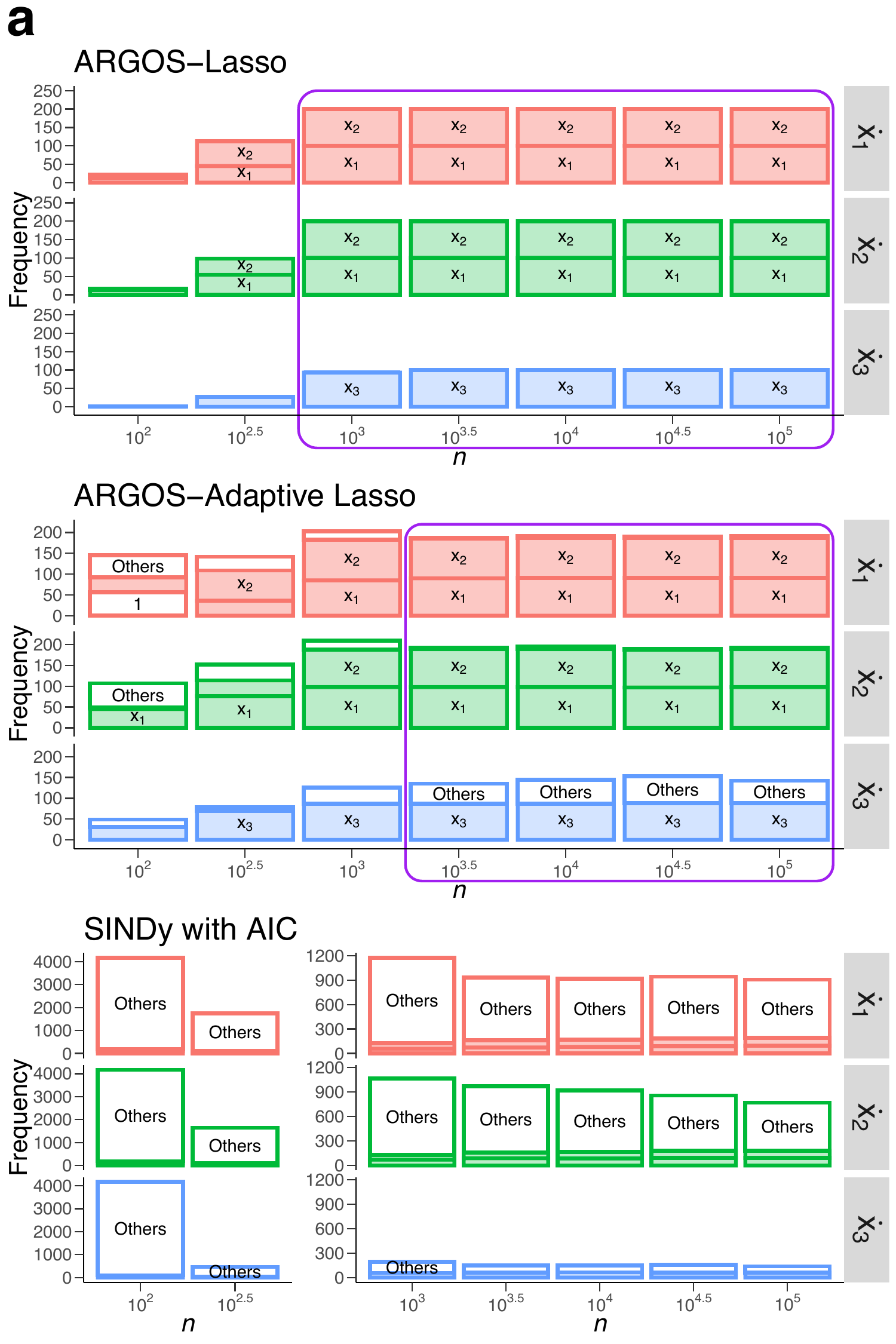}
\end{minipage}
\begin{minipage}[c]{0.495\linewidth}
\includegraphics[width=1\linewidth]{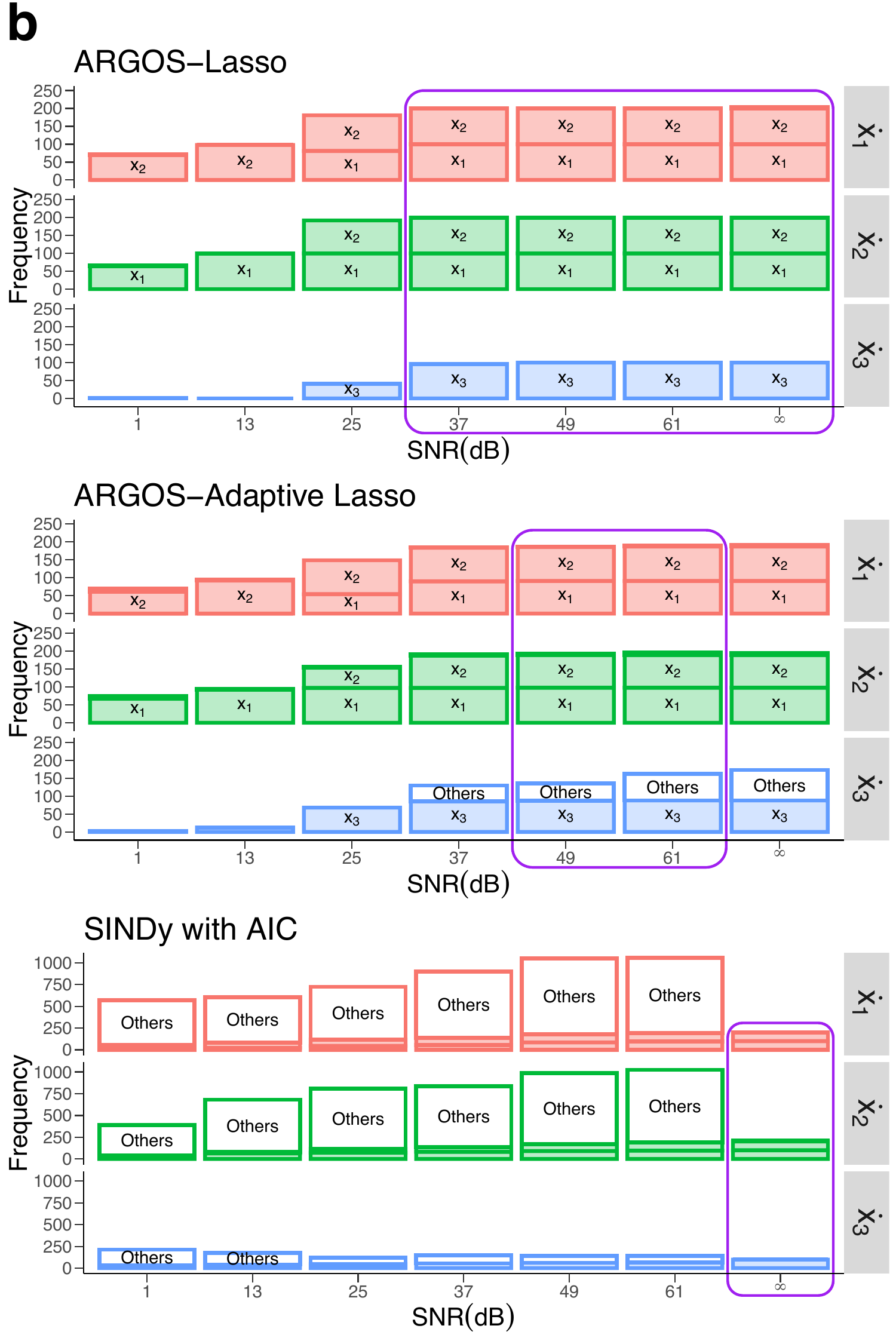}
\end{minipage}
\includegraphics[width=0.3\textwidth]{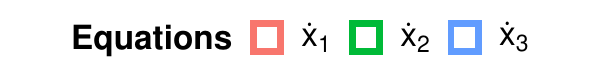}
\caption{\textbf{Frequency of identified variables for the three-dimensional linear system across algorithms.}
Colors correspond to each governing equation; filled boxes indicate correctly identified variables, while white boxes denote erroneous terms.
Panels show the frequency of identified variables for data sets with (\textbf{a}) increasing $n$ (SNR = 49 dB), and (\textbf{b}) SNR ($n=5000$).
Purple-bordered regions demarcate model discovery above 80\%.
}
\label{fig:linear3d_stacked}
\end{figure*}
For $x_1(t)$, $x_2(t)$, and $x_3(t)$, we again developed a random uniform distribution containing 100 values between $[10^{-1}, 10^3]$.

Figure~\ref{fig:linear3d_stacked} demonstrates the efficacy of employing the lasso within our framework, as the method more consistently identified the three-dimensional linear system than the other two algorithms studied here.
With greater than 1000 observations and medium SNR, our approach accurately represented these simple dynamics.
We also found that, with low SNR and less than 300 observations, our method discovered models that were overly sparse and did not fully represent the dynamics, while SINDy with AIC often failed to discover a parsimonious representation of the system by identifying dense models with superfluous terms.

\section{First-order nonlinear systems}
\label{supp_sect:first_order_nonlinear}
\subsection*{Two-dimensional damped oscillator with cubic dynamics}
\label{subsupp_sect:cubic2d_system}
We examined a two-dimensional system with cubic dynamics as~\cite{bruntonDiscoveringGoverningEquations2016}
\begin{equation}
\begin{aligned}
\label{eq:cubic2D_system}
    \dot{x}_1 &= -0.1x_1^{3} + 2x_2^{3}, \\
    \dot{x}_2 &= -2x_1^{3} - 0.1x_2^{3} .
\end{aligned}
\end{equation}
\begin{figure*}[!htp]
\centering
\begin{minipage}[c]{0.495\linewidth}
\includegraphics[width=1\linewidth]{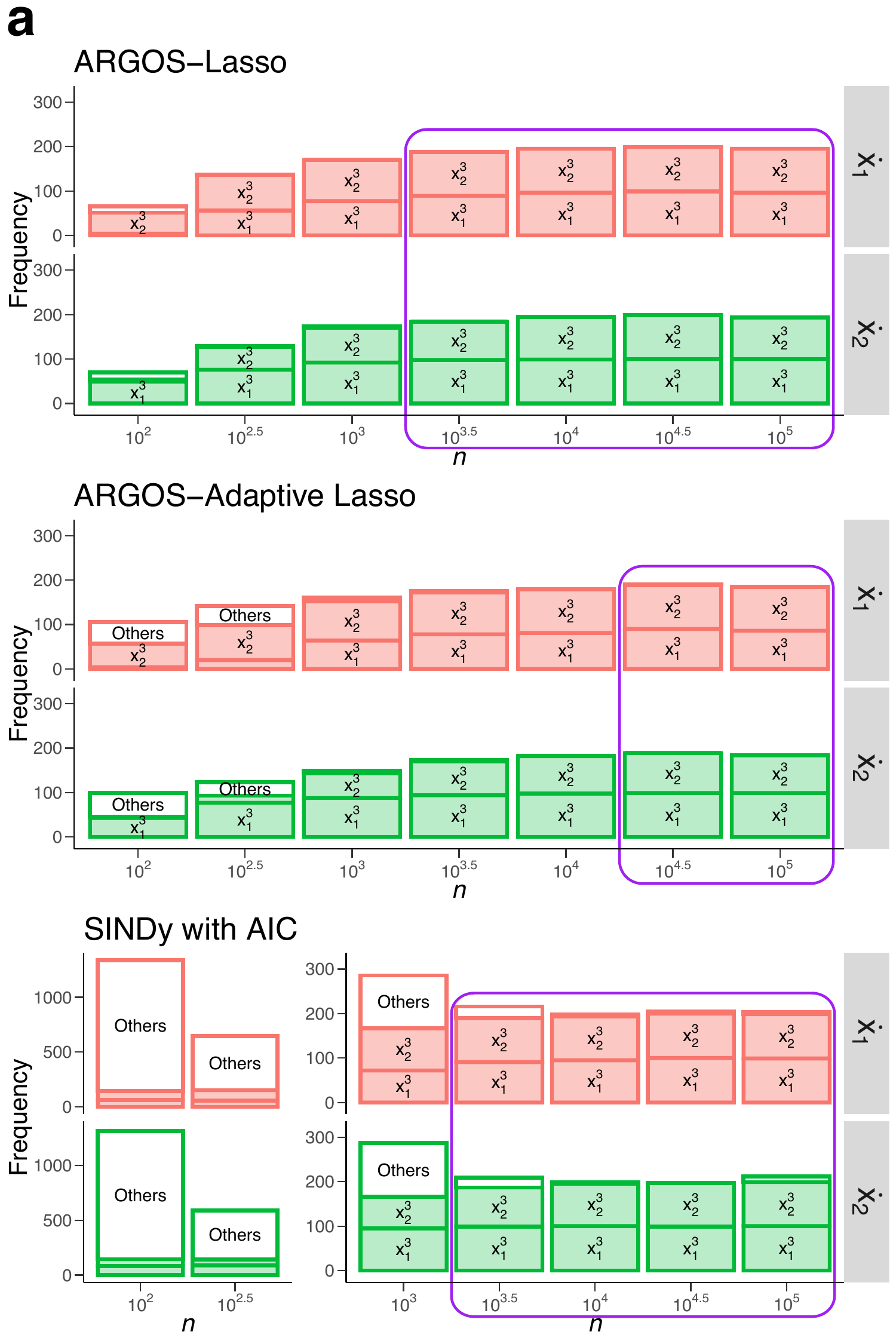}\end{minipage}
\begin{minipage}[c]{0.495\linewidth}
\includegraphics[width=1\linewidth]{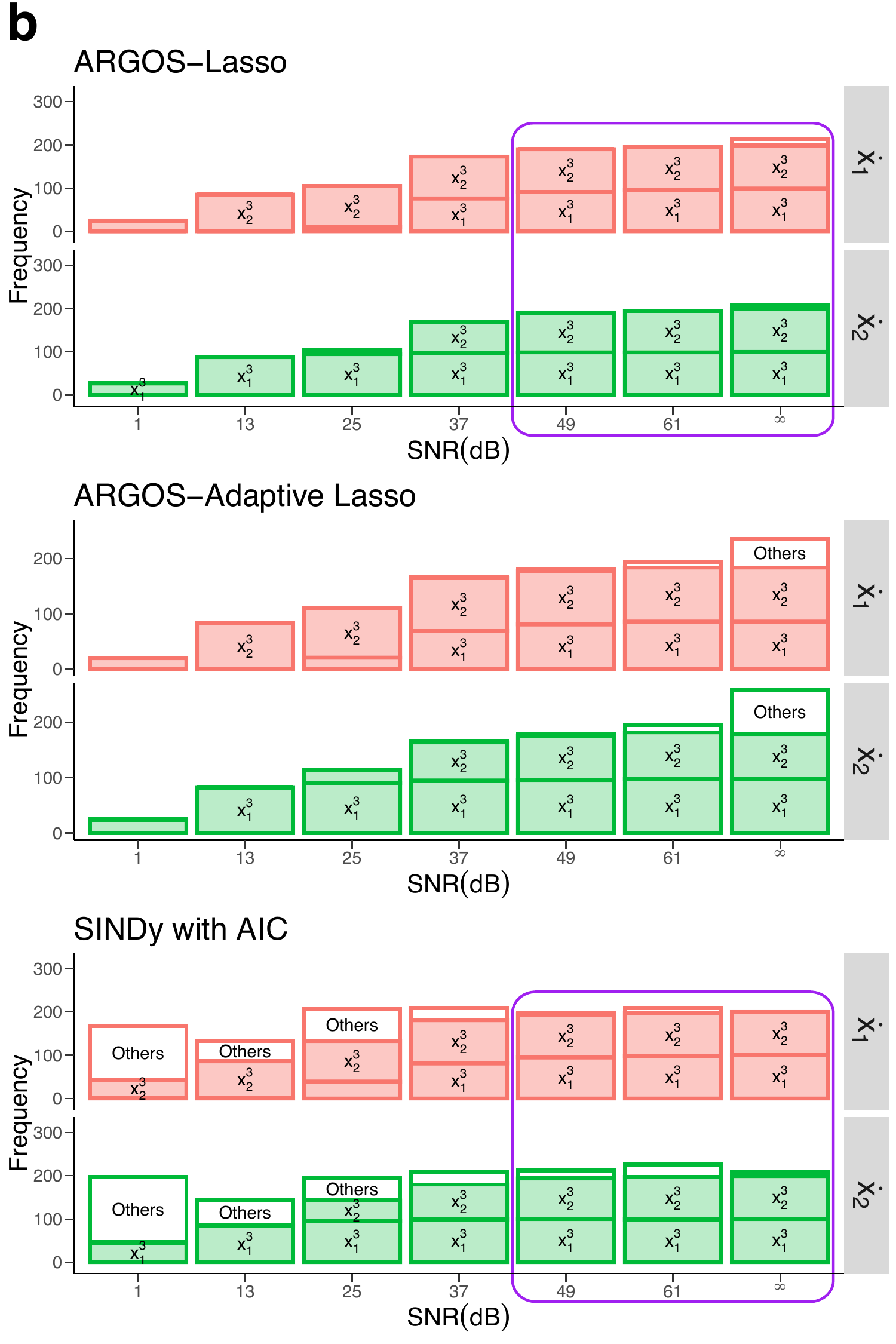}
\end{minipage}
\includegraphics[width=0.3\textwidth]{Figures/legend_2d.pdf}
\caption{\textbf{Frequency of identified variables for the two-dimensional damped oscillator with cubic dynamics across algorithms.}
Colors correspond to each governing equation; filled boxes indicate correctly identified variables, while white boxes denote erroneous terms.
Panels show the frequency of identified variables for data sets with (\textbf{a}) increasing $n$ (SNR = 49 dB), and (\textbf{b}) SNR ($n=5000$).
Purple-bordered regions demarcate model discovery above 80\%.
}
\label{fig:cubic2d_stacked}
\end{figure*}
For $x_1(t)$ and $x_2(t)$, we generated a random uniform distribution containing 100 values between $[-2, 2]$.

Figure~\ref{fig:cubic2d_stacked} further demonstrates the effectiveness of the lasso algorithm for identifying the two-dimensional damped harmonic oscillator with cubic dynamics.
Here, SINDy with AIC performed model discovery with similar accuracy to our approach, while both the lasso and SINDy with AIC ultimately outperformed the adaptive lasso.

\subsection*{Lotka-Volterra system}
\label{subsupp_sect:lotka_volterra_system}
The Lotka-Volterra system is described by two first-order nonlinear differential equations commonly used to depict the interaction dynamics between two species in biological systems, with one being the predator and the other the prey~\citep{lotkaContributionTheoryPeriodic1910}.
The predator-prey equations are represented as
\begin{equation}
\begin{aligned}
\label{eq:lotka_volterra_ch3}
    \dot{x}_1 &= \alpha x_1 - \zeta x_1x_2, \\
    \dot{x}_2 &= \delta x_1x_2 - \gamma x_2,
\end{aligned}
\end{equation}
where $\alpha=1$ represents the prey birth rate and $\delta=-1$ is the predator death rate, and $\zeta=-1$ and $\gamma=1$ are the interaction parameters~\citep{naozukaSINDySAFrameworkEnhancing2022}.
Since the population cannot be negative, we used a random uniform distribution with 100 positive values between $[1, 10]$ for the initial conditions of both $x_1(t)$ and $x_2(t)$.

\begin{figure*}[!htp]
\centering
\begin{minipage}[c]{0.495\linewidth}
\includegraphics[width=1\linewidth]{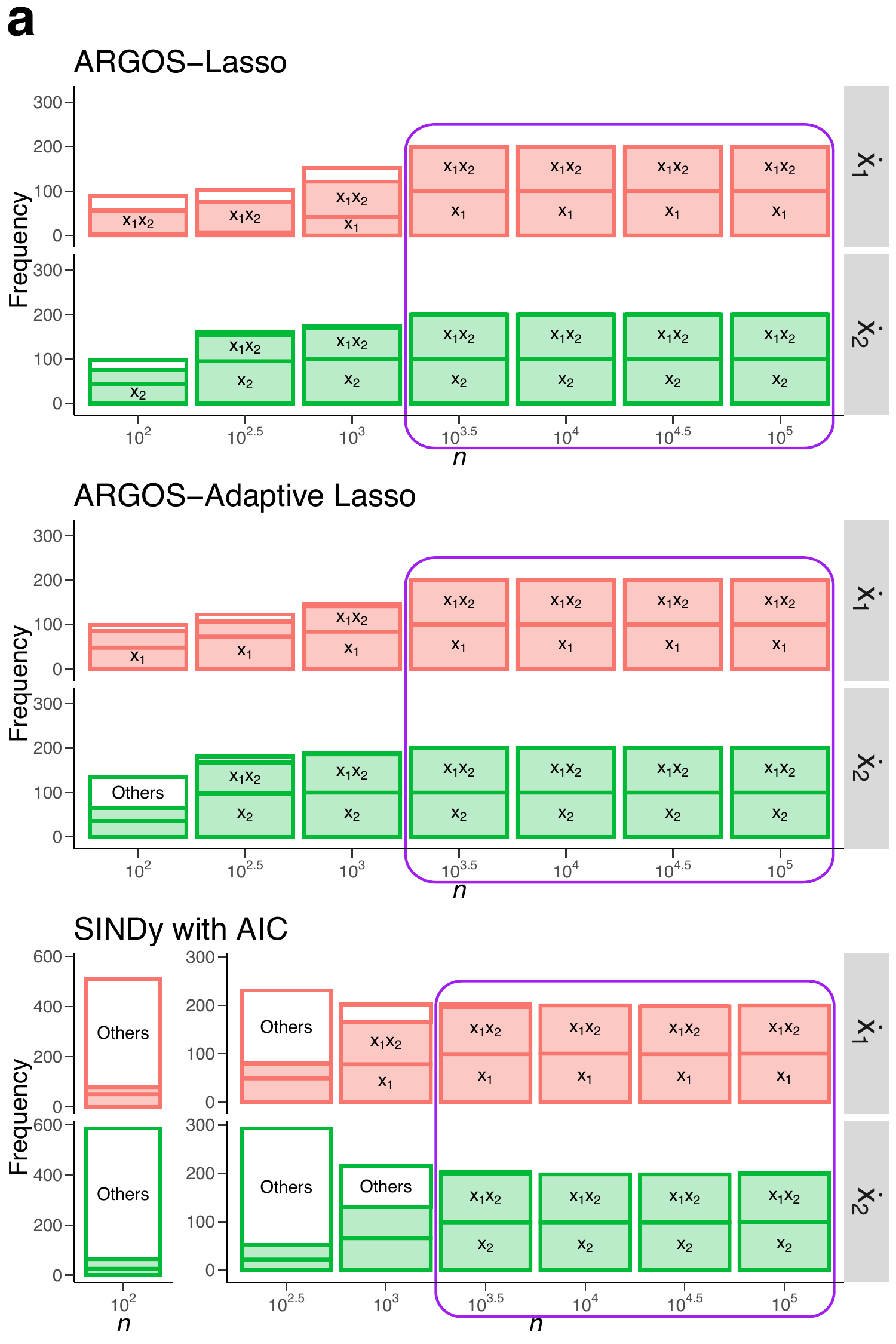}
\end{minipage}
\begin{minipage}[c]{0.495\linewidth}
\includegraphics[width=1\linewidth]{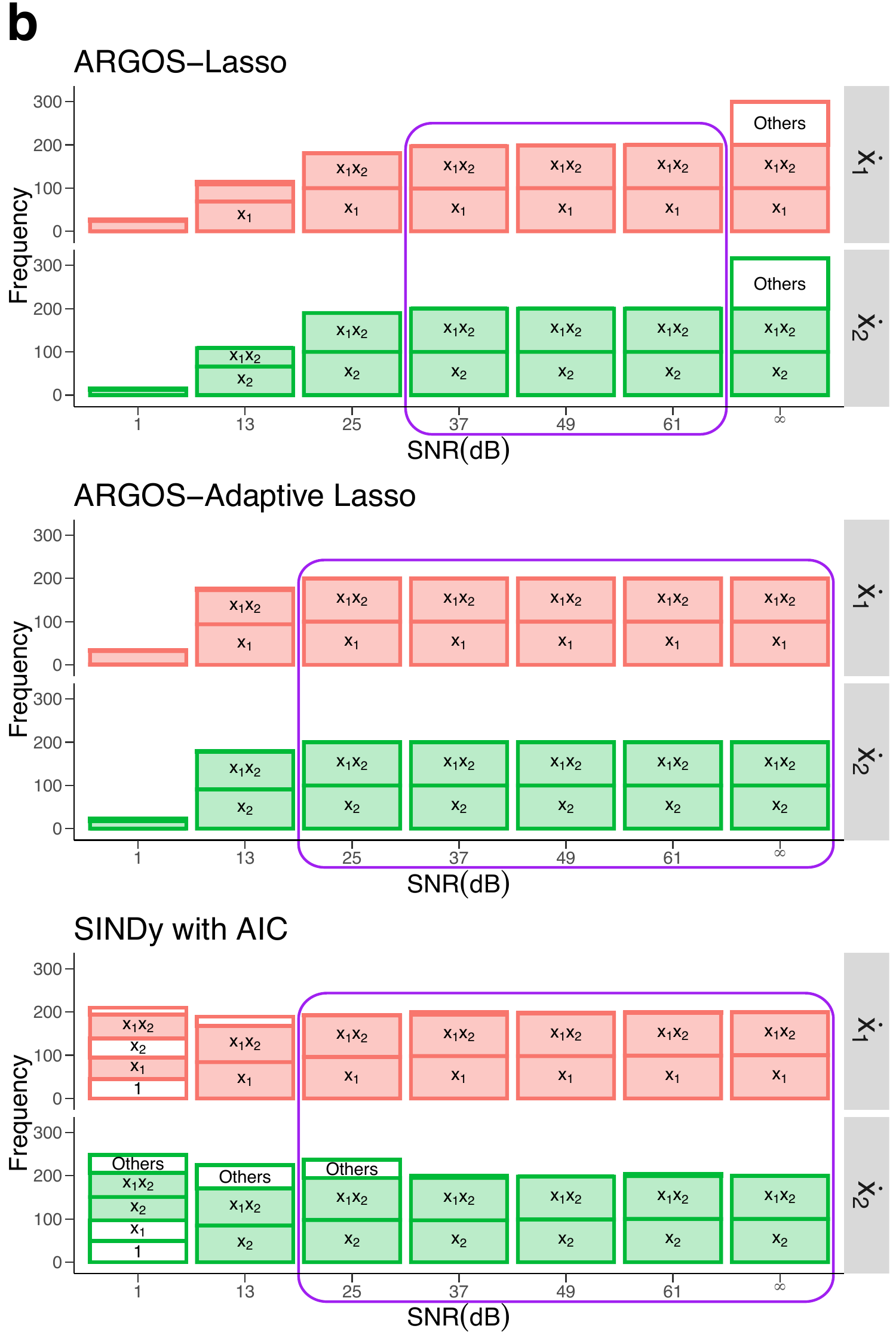}
\end{minipage}
\includegraphics[width=0.3\textwidth]{Figures/legend_2d.pdf}
\caption{\textbf{Frequency of identified variables for the Lotka-Volterra system across algorithms.}
Colors correspond to each governing equation; filled boxes indicate correctly identified variables, while white boxes denote erroneous terms.
Panels show the frequency of identified variables for data sets with (\textbf{a}) increasing $n$ (SNR = 49 dB), and (\textbf{b}) SNR ($n=5000$).
Purple-bordered regions demarcate model discovery above 80\%.
}
\label{fig:lotka_volterra_stacked}
\end{figure*}

Figure~\ref{fig:lotka_volterra_stacked} illustrates that, as $n$ and SNR increased, we most consistently identified the true governing terms of the equations using the adaptive lasso within our framework.
In contrast, SINDy with AIC tended to discover numerous erroneous terms when data contained fewer than 3000 observations and low SNR.

\subsection*{Rossler system}
\label{subsupp_sect:rossler}
We examined the Rossler system, a three-dimensional chaotic system represented as
\begin{equation}
\begin{aligned}
\label{eq:rossler_system}
    \dot{x}_1 &= -x_2-x_3, \\
    \dot{x}_2 &= x_1 + ax_2, \\
    \dot{x}_3 &= b +x_3(x_1-c),
\end{aligned}
\end{equation}
where $a = 0.2$, $b = 0.2$, and $c = 5.7$~\cite{tranExactRecoveryChaotic2017}.
For $x_1(t)$, $x_2(t)$, and $x_3(t)$, we generated a random uniform distribution containing 100 values between $[-10, 10]$, $[-10, 10]$, and $[0, 20]$.
\begin{figure*}[!htp]
\centering
\begin{minipage}[c]{0.495\linewidth}
\includegraphics[width=1\linewidth]{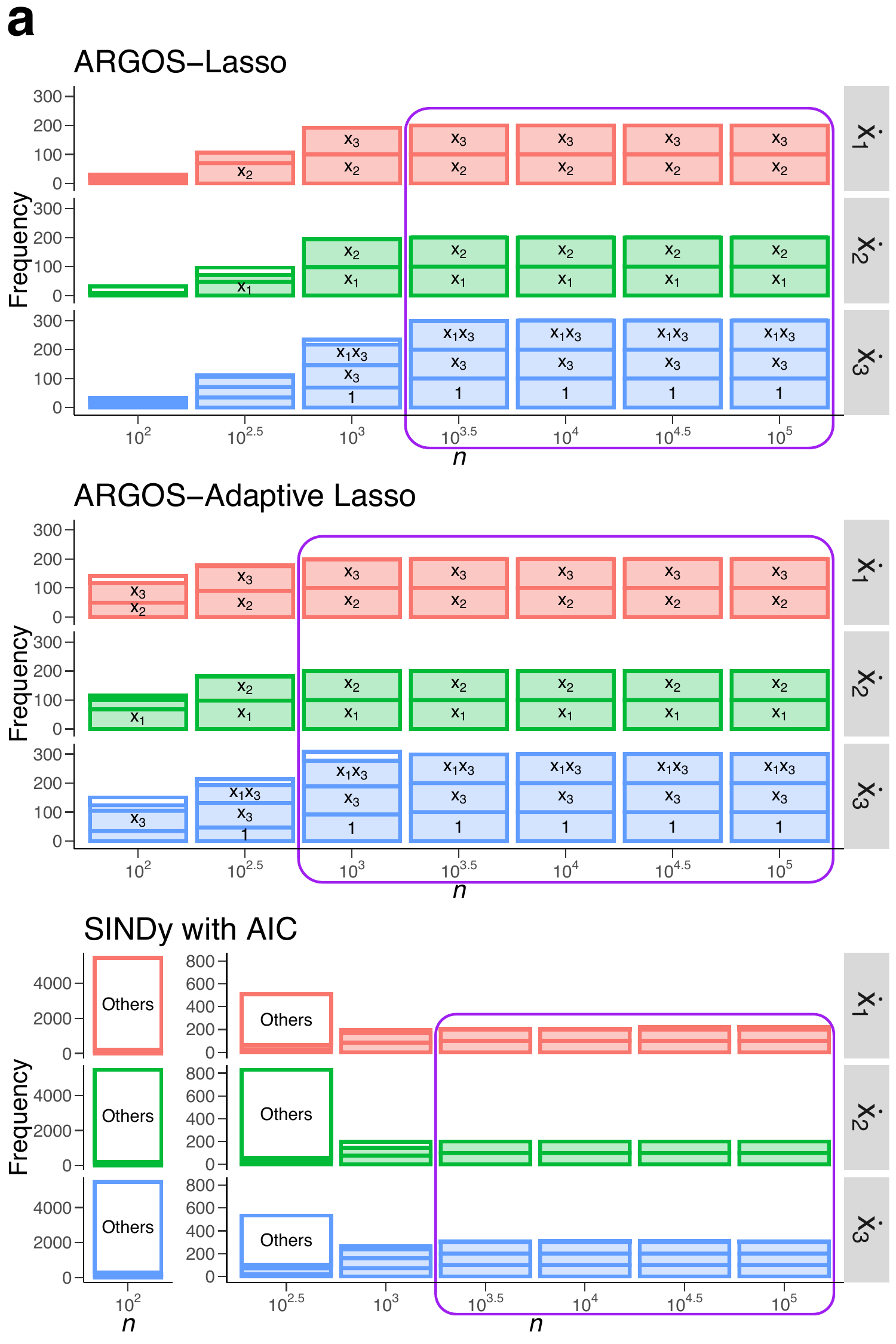}
\end{minipage}
\begin{minipage}[c]{0.495\linewidth}
\includegraphics[width=1\linewidth]{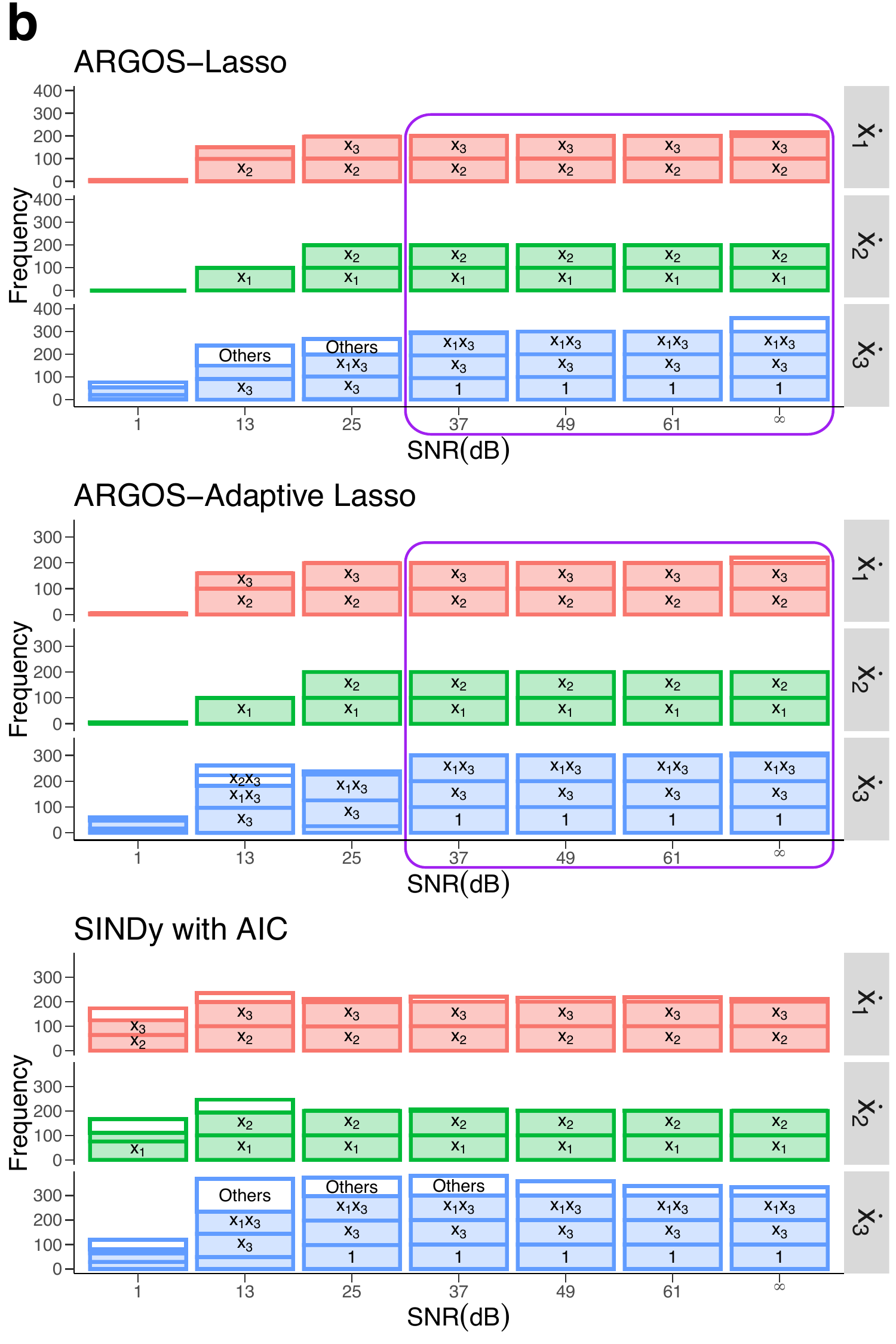}
\end{minipage}
\includegraphics[width=0.3\textwidth]{legend_3d.pdf}
\caption{\textbf{Frequency of identified variables for the Rossler system across algorithms.}
Colors correspond to each governing equation; filled boxes indicate correctly identified variables, while white boxes denote erroneous terms.
Panels show the frequency of identified variables for data sets with (\textbf{a}) increasing $n$ (SNR = 49 dB), and (\textbf{b}) SNR ($n=5000$).
Purple-bordered regions demarcate model discovery above 80\%.
}
\label{fig:rossler_stacked}
\end{figure*}

Figure~\ref{fig:rossler_stacked} demonstrates the effectiveness of our approach in accurately representing the Rossler system, provided that sufficient data is available.
Here, our method consistently identified the underlying dynamics while SINDy with AIC failed to surpass 80\% success for any SNR value, emphasizing the limitations of the sequential thresholding procedure.

\subsection*{Lorenz system}
\label{subsupp_sect:Lorenz}
We examined the Lorenz chaotic system, a low-dimensional nonlinear structure originally a simple model for atmospheric convection.
The Lorenz systems are modeled using the following equations:
\begin{equation}
\begin{aligned}
\label{eq:lorenz_system}
    \dot{x}_1 &= \sigma(x_2-x_1), \\
    \dot{x}_2 &= x_1(\rho - x_3) -x_2, \\
    \dot{x}_3 &= x_1x_2 - \zeta x_3,
\end{aligned}
\end{equation}
with the values of the original parameters $\sigma=10$, $\rho=28$, and $\zeta = 8/3$~\cite{bruntonDiscoveringGoverningEquations2016}.
For $x_1(t)$, $x_2(t)$, and $x_3(t)$, we developed a random uniform distribution containing 100 values between $[-15, 15]$, $[-15, 15]$, and $[10, 40]$.
The \nameref{sect:results} section provides more detail regarding each method's performance in discovering the Lorenz system.

\section{Second-order nonlinear systems}
\label{supp_sect:second_order_nonlinear}
\subsection*{Van der Pol oscillator}
\label{supp_sect:vdp_oscillator}
We examined the Van der Pol oscillator, introduced in 1922 as a nonlinear circuit model with a triode tube, represented as
\begin{equation}
\begin{aligned}
\label{eq:vdpol_oscillator}
    \dot{x}_1 &= x_2, \\
    \dot{x}_2 &= \mu(1-x_1^2)x_2 - x_1,
\end{aligned}
\end{equation}
where $\mu=1.2$ controls the nonlinear damping level of the system~\cite{cortiellaSparseIdentificationNonlinear2021}.
For $x_1(t)$ and $x_2(t)$, we developed a random uniform distribution containing 100 values between $[-4, 4]$.
\begin{figure*}[!htp]
\centering
\begin{minipage}[c]{0.495\linewidth}
\includegraphics[width=1\linewidth]{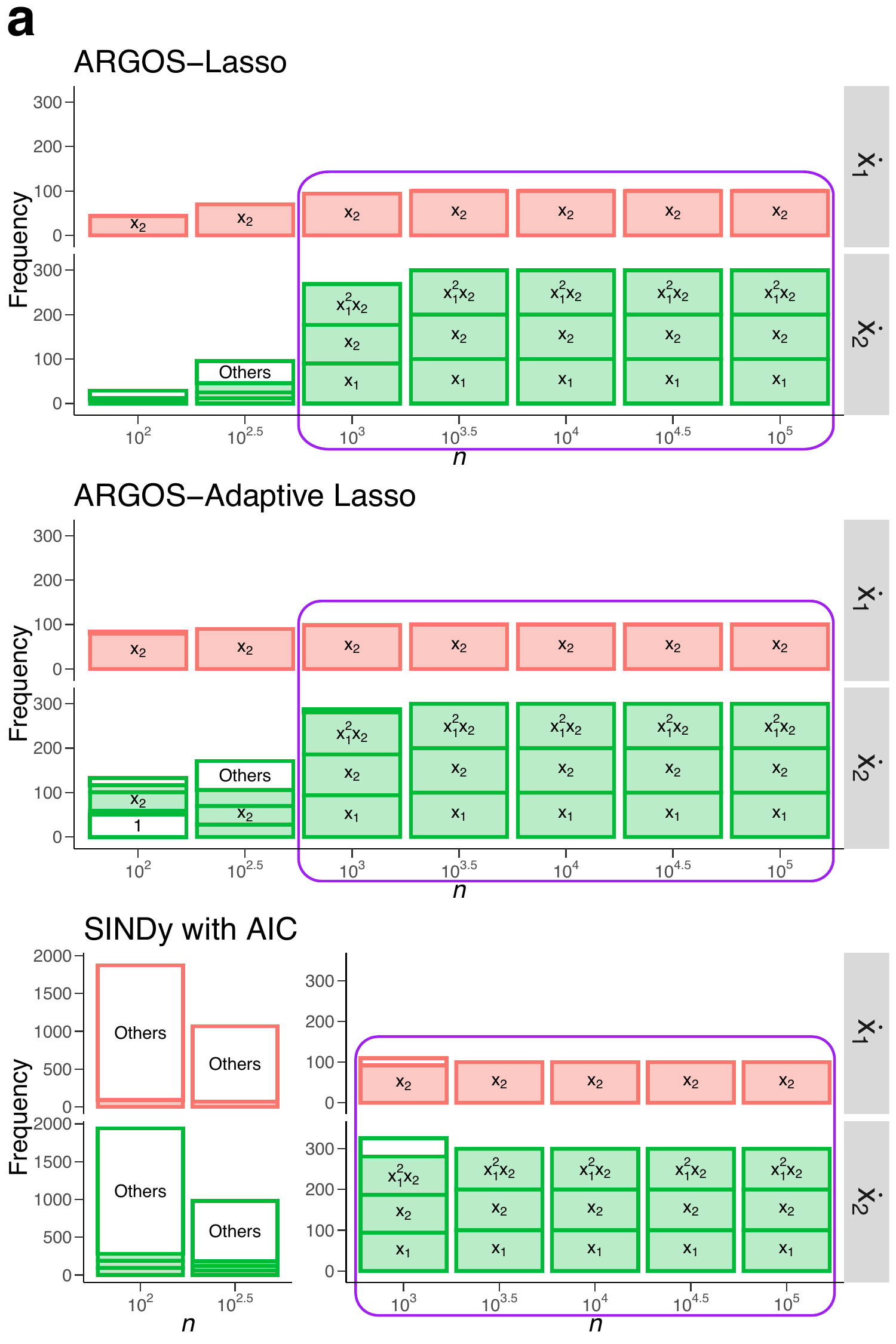}
\end{minipage}
\begin{minipage}[c]{0.495\linewidth}
\includegraphics[width=1\linewidth]{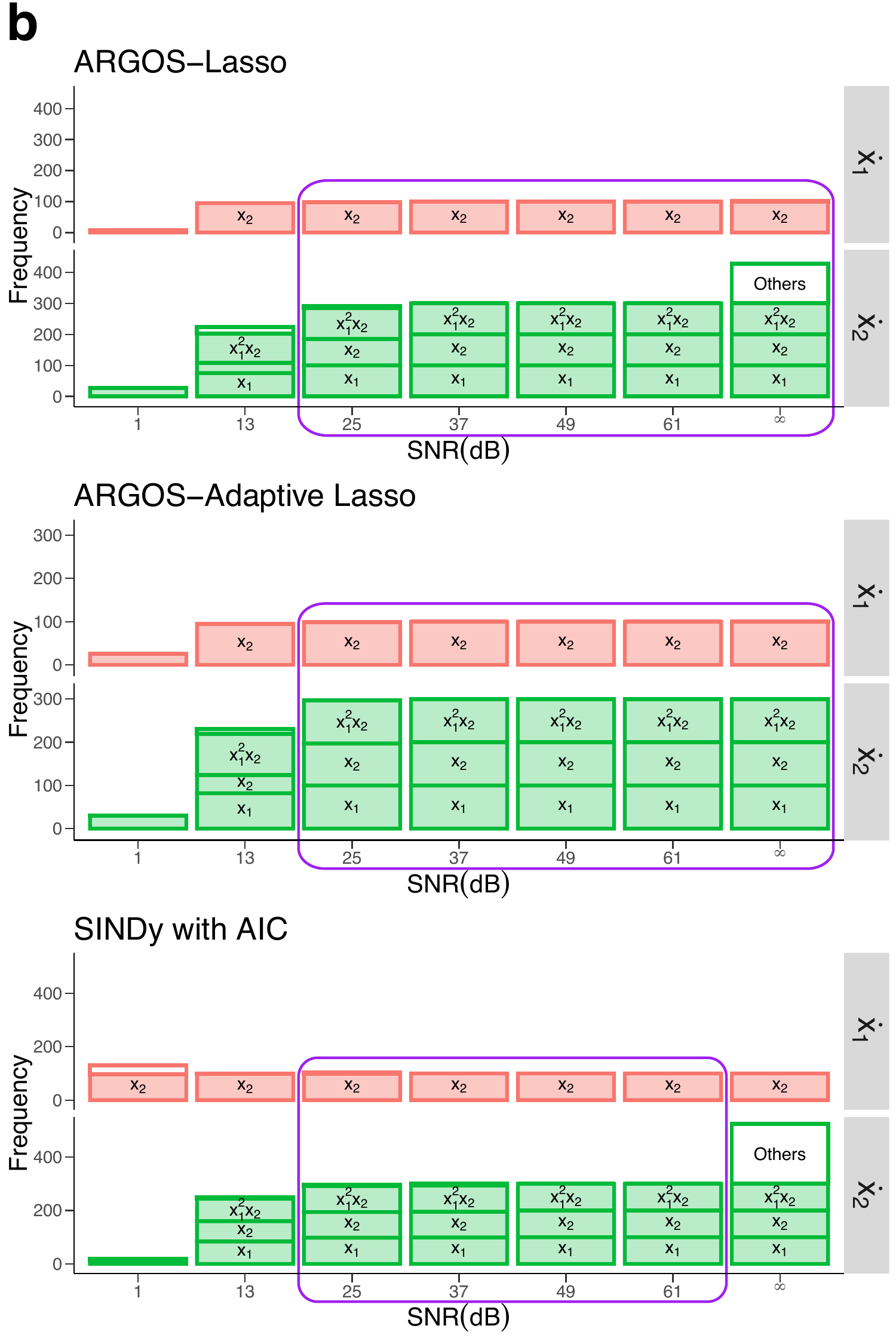}
\end{minipage}
\includegraphics[width=0.3\textwidth]{Figures/legend_2d.pdf}
\caption{\textbf{Frequency of identified variables for the Van der Pol oscillator across algorithms.}
Colors correspond to each governing equation; filled boxes indicate correctly identified variables, while white boxes denote erroneous terms.
Panels show the frequency of identified variables for data sets with (\textbf{a}) increasing $n$ (SNR = 49 dB), and (\textbf{b}) SNR ($n=5000$).
Purple-bordered regions demarcate model discovery above 80\%.
}
\label{fig:vdp_stacked}
\end{figure*}

Figure~\ref{fig:vdp_stacked} further illustrates the tendencies of each algorithm when faced with a limited number of observations and low SNR.
Under these conditions, our method developed overly sparse models, while SINDy with AIC produced dense models that did not accurately represent the Van der Pol oscillator.
However, as $n$ and SNR increased, our approach demonstrated a marked improvement in accurately discovering the underlying equations of the oscillator.

\subsection*{Duffing oscillator}
\label{subsupp_sect:duffing}
We examined the Duffing oscillator as an alternative cubic nonlinear system that can represent chaos.
The Duffing oscillator models a spring-damper-mass system that contains a spring with a restoring force of $f(\zeta) = -\kappa\zeta - \epsilon\zeta^3$, where $\epsilon > 0$ represents a hard spring~\cite{cortiellaSparseIdentificationNonlinear2021}.
However, when $\epsilon < 0$, it represents a soft spring and is given by
\begin{equation}
\begin{aligned}
\label{eq:duffing_system_initial}
    \ddot{\zeta}_1 + \gamma\dot{\zeta} + (\kappa + \epsilon\zeta^2)\zeta = 0.
\end{aligned}
\end{equation}
We converted $x = \zeta$ and $y = \dot{\zeta}$ and transformed Eq.~\eqref{eq:duffing_system_initial} to
\begin{equation}
\begin{aligned}
\label{eq:duffing_system}
    \dot{x}_1 &= x_2, \\
    \dot{x}_2 &= -\gamma x_2 - \kappa x_1 - \epsilon x_1^3.
\end{aligned}
\end{equation}
Here, we used parameter values for which the Duffing oscillator does not represent chaotic behavior: $\kappa = 1$, $\gamma = 1$, and $\epsilon = 5$ ~\cite{cortiellaSparseIdentificationNonlinear2021}.
For $x_1(t)$ and $x_2(t)$, we developed a random uniform distribution containing 100 values between $[-2, 2]$, $[-6, 6]$.
\begin{figure*}[!htp]
\centering
\begin{minipage}[c]{0.495\linewidth}
\includegraphics[width=1\linewidth]{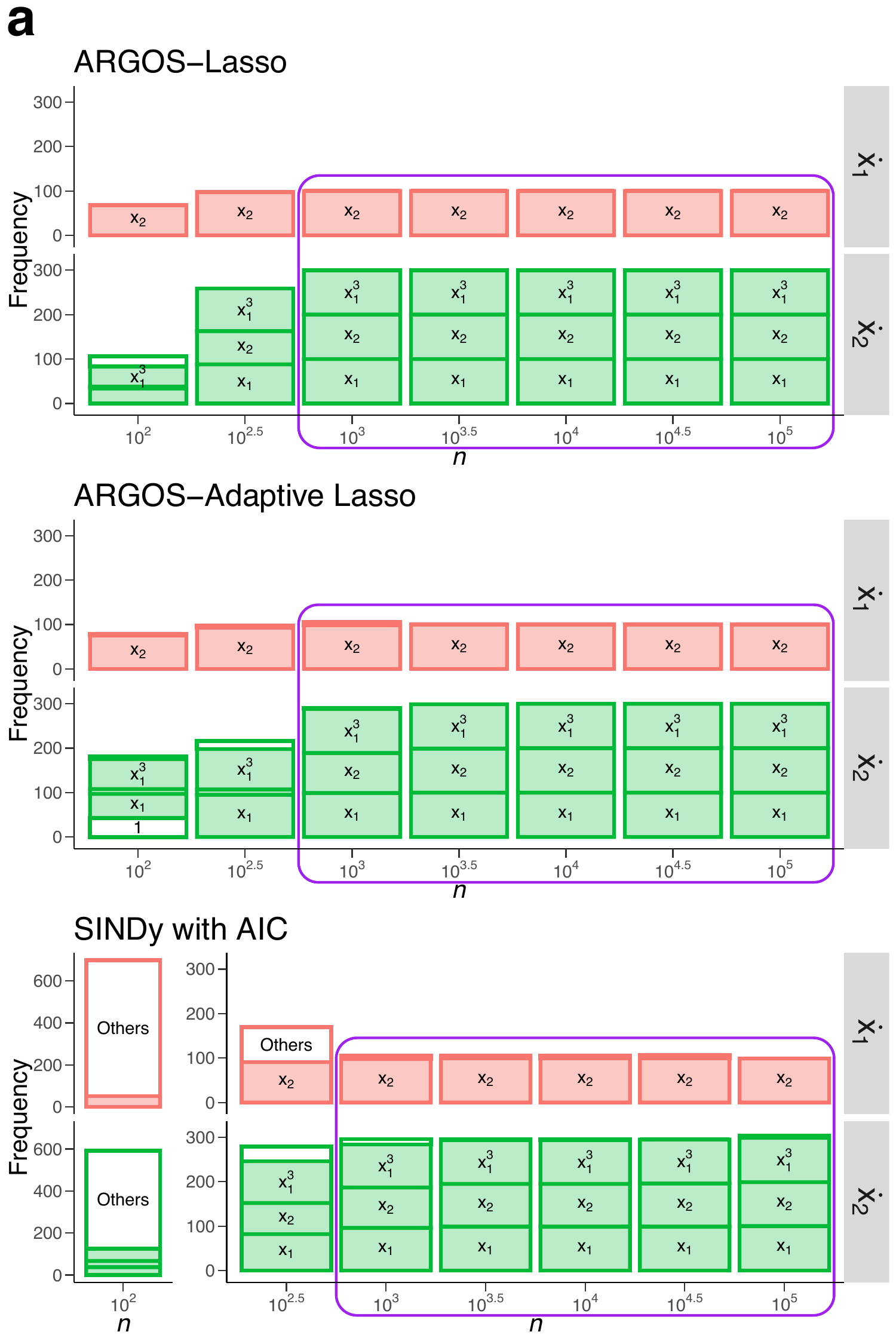}
\end{minipage}
\begin{minipage}[c]{0.495\linewidth}
\includegraphics[width=1\linewidth]{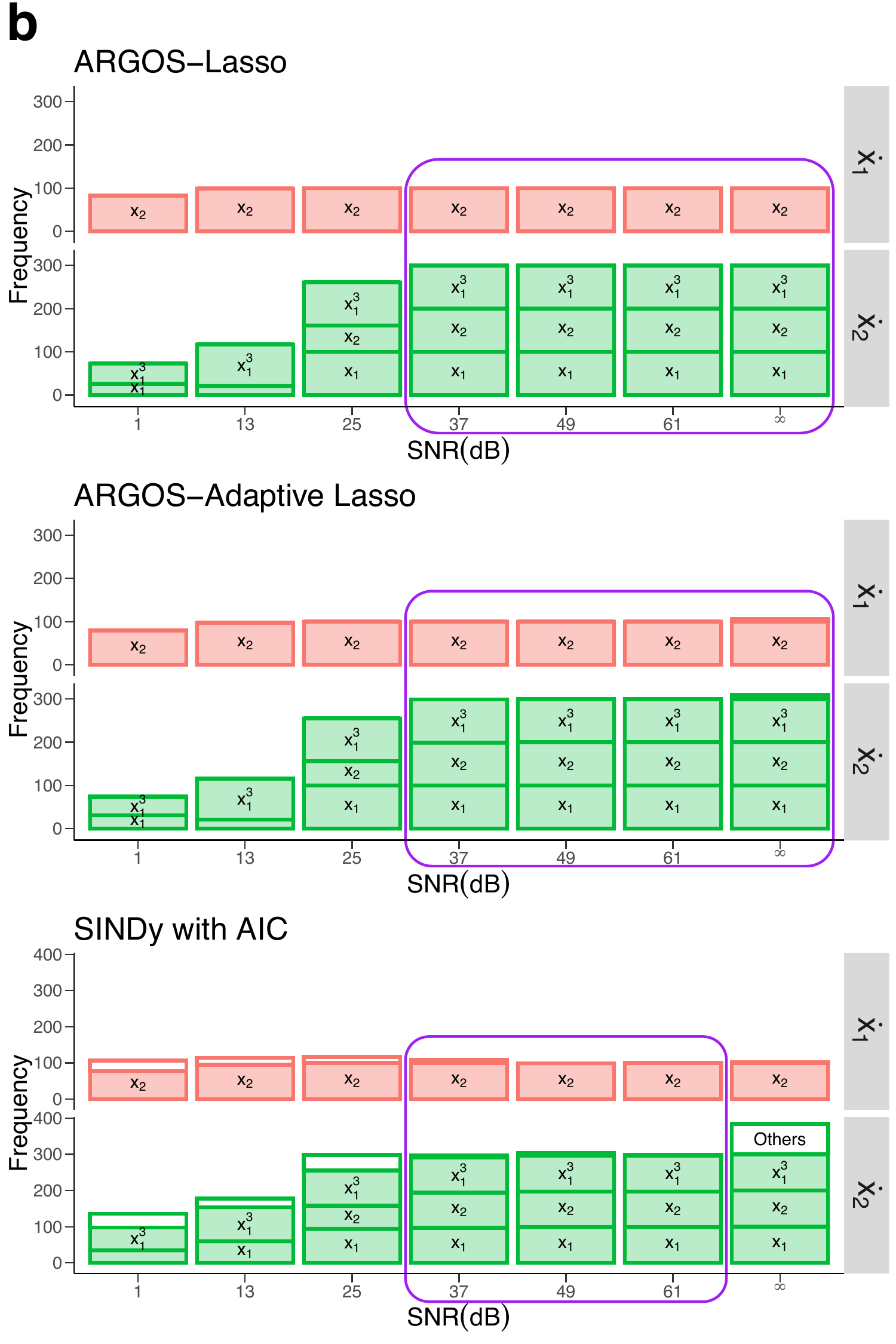}
\end{minipage}
\includegraphics[width=0.3\textwidth]{Figures/legend_2d.pdf}
\caption{\textbf{Frequency of identified variables for the Duffing oscillator across algorithms.}
Colors correspond to each governing equation; filled boxes indicate correctly identified variables, while white boxes denote erroneous terms.
Panels show the frequency of identified variables for data sets with (\textbf{a}) increasing $n$ (SNR = 49 dB), and (\textbf{b}) SNR ($n=5000$).
Purple-bordered regions demarcate model discovery above 80\%.
}
\label{fig:duffing_stacked}
\end{figure*}

Figure~\ref{fig:duffing_stacked} shows that our method consistently represented the Duffing oscillator with high accuracy as $n$ and SNR increased.
In this example, with less than 1000 observations and low SNR, our approach developed overly sparse models that inadequately captured the dynamics of the system, while SINDy with AIC again developed dense models that misrepresented the dynamics.

\clearpage
\section{Algorithms}
\label{supp_sect:algorithms}
\begin{algorithm*}[!htp]
\caption{Automatic Savitzky-Golay Filter}
\label{al:SG}
\begin{algorithmic}[1]
\Require $\mathbf{\Tilde{X}}\in \reals^{n\times m}$, $\ dt$.
\Ensure Savitzky-Golay optimally smoothed $\mathbf{X}$ and $\mathbf{\dot{X}}$.
\\

determine lower and upper bounds of (odd) window length $l$:

$l_{min}= 13$,

$l_{max} = \max\left(13,\ \min\left(n - \left(n - 1\right)\bmod2, 101\right )\ \right)$;\\

build $L = (l_{min},\dots, l_{max})$; 

\color{blue}
\LeftComment{$v=$ degree of derivative}
\normalcolor
\For{$j=1,\dots,m$}

$l^{\ast} = \argmin_{L}\|SG(\mathbf{\Tilde{x}}_j, o = 4, l = L, v = 0, dt) - \mathbf{\Tilde{x}}_j\|^2_2$,

$\mathbf{x}_j = SG(\mathbf{\Tilde{x}}_j, o = 4, l=l^{\ast}, v=0, dt)$,

$\mathbf{\dot{x}}_j = SG(\mathbf{\Tilde{x}}_j, o = 4, l=l^{\ast}, v=1, dt)$;
\EndFor\\

consolidate $\mathbf{X},\mathbf{\dot{X}}\in\reals^{n\times m}$ with each $\mathbf{x}_j$ and $\mathbf{\dot{x}}_j$, respectively.
\end{algorithmic}
\end{algorithm*}
\bigskip
\begin{algorithm}[!htp]
\caption{Automatic Regression for Governing Equations (ARGOS)}
\label{al:ARGOS}

\begin{algorithmic}[1]
\Require $\mathbf{X}\in \reals^{n\times m}$, $\mathbf{\dot{x}}_j\in \reals^{n}$, $d$, $\alpha = 0.05$.
\color{blue}
\LeftComment{STEP ONE: Initial design matrix}
\normalcolor\\
$p^{(0)}=\binom{m+d}{d}$;\\

create $\mathbf{\Theta}^{(0)}(\mathbf{X}) \in \reals^{n\times p^{(0)}}$ with basis functions up to order $d$ of the columns of $\mathbf{X}$;

\color{blue}
\LeftComment{STEP TWO: Trim design matrix}

\LeftComment{Variable selection with the lasso or adaptive lasso}
\LeftComment{$\lambda^\ast$ : Optimal $\lambda$ from 10-fold cross-validation}
\LeftComment{lasso: $w=1$}
\LeftComment{adaptive lasso: $w=$ ridge regression coefficients}\normalcolor\\
$\begin{aligned}
\hat{\beta}^{(0)} = \argmin_{\beta} \left\|\mathbf{\dot{x}}_{j} -\mathbf{\Theta}^{(0)}(\mathbf{X})\beta\right\|_2^2 + \lambda^\ast \sum_{k=1}^{p^{(0)}}w_{k}\abs{\beta_{k}}
\end{aligned}$;\label{algo_step:initial_pe_sparse_regression}\\

extract $\mathbf{\Theta}^{(1)}(\mathbf{X})$ to contain columns of $\mathbf{\Theta}^{(0)}(\mathbf{X})$ up to the largest order $d^{(1)}$ of the selected variables in $\hat{\beta}^{(0)}$;

\color{blue}
\LeftComment{STEP THREE: Final point estimates}
\LeftComment{Repeat sparse regression algorithm from STEP TWO}\normalcolor\\

$p^{(1)}=\binom{m+d^{(1)}}{d^{(1)}}$;\\

$\begin{aligned}
\hat{\beta}^{(1)}
  &= \argmin_{\beta} \left\|\mathbf{\dot{x}}_{j} -\mathbf{\Theta}^{(1)}(\mathbf{X})\beta\right\|_2^2 + \lambda^\ast \sum_{k=1}^{p^{(1)}}w_{k}\abs{\beta_{k}};
\end{aligned}$\label{algo_step:secondary_pe_sparse_regression}
\color{blue}
\LeftComment{Apply threshold values}
\normalcolor\\
$\eta = [10^{-8},10^{-7},\dotsc, 10^{1}]$;

\For{$i=1,\dots, \mathbf{card}(\eta)$}
\color{blue}
\LeftComment{Ordinary least squares regression (OLS) estimate after variable selection}\normalcolor

$\begin{aligned}
\hat{\beta}^{\text{OLS}}[i]
  &= \argmin_{\beta_{\mathcal{K}_i}} \left\|\mathbf{\dot{x}}_{j} -\mathbf{\Theta}_{\mathcal{K}_i}^{(1)}(\mathbf{X})\beta_{\mathcal{K}_i}\right\|_2^2 \ \text{where} \ \mathcal{K}_i = \{k:\abs{\hat{\beta}_{k}^{(1)}} \geq \eta_i\},\\
\end{aligned}$

$\text{BIC}_i = \text{BIC}(\hat{\beta}^{\text{OLS}}[i])$;
\EndFor\\
$\hat{\beta} =  \left\{ \hat{\beta}^{\text{OLS}}[i]\middle\vert i:  \argmin(\text{BIC}) \right\};$\label{algo_step:final_point_estimate}

\color{blue}
\LeftComment{STEP FOUR: Bootstrap estimates for confidence intervals}

\LeftComment{$B = 2000$ bootstrap samples}
\normalcolor\\

bootstrap Statements~\ref{algo_step:secondary_pe_sparse_regression}~--~\ref{algo_step:final_point_estimate} to approximate confidence interval bounds: $CI_\text{lo} = [B\alpha/2]$ and  $CI_\text{up} = B-CI_\text{lo}+1$;\\

construct bootstrap confidence intervals for $\hat{\beta}$:
$\begin{aligned}
\hat{\beta}_{k} \in \left[\hat{\beta}_{k}^{\text{OLS}\{CI_\text{lo}\}}, \hat{\beta}_{k}^{\text{OLS}\{CI_\text{up}\}} \right] \text{, and }
0<\hat{\beta}_{k}^{\text{OLS}\{CI_\text{lo}\}} \text{ or } 0>\hat{\beta}_{k}^{\text{OLS}\{CI_\text{up}\}}
\end{aligned}$.
\end{algorithmic}
\end{algorithm}
\newpage

\end{document}